\documentclass[runningheads]{llncs}
\usepackage{ifthen,todonotes}
 \newcommand{\otc}[1]{\textcolor{black}{#1}}

\newcommand{\carola}[1]{\textcolor{black}{#1}}

  \newcommand{\olivier}[1]{\textcolor{black}{#1}}

\makeatletter
\newcommand*{\inlineequation}[2][]{%
  \begingroup
    \refstepcounter{equation}%
    \ifx\\#1\\%
    \else
      \label{#1}%
    \fi
    \relpenalty=10000 %
    \binoppenalty=10000 %
    \ensuremath{%
      #2%
    }%
    ~\@eqnnum
  \endgroup
}
\makeatother

\renewcommand{\epsilon}{\varepsilon}

\usepackage[utf8]{inputenc}

\usepackage{amssymb,amsmath,url,rotating,ifthen,algorithm,algorithmicx,epsfig,multirow,array,color,multicol,graphicx,float,hhline,calc,algpseudocode}
\usepackage{todonotes} 
\usepackage{bold-extra,ulem}
\usepackage{enumitem}
\setlistdepth{9}
\setlist[itemize,1]{label=$\bullet$}
\setlist[itemize,2]{label=$-$}
\setlist[itemize,3]{label=$\ast$}
\setlist[itemize,4]{label=$\cdot$}
\setlist[itemize,5]{label=$\diamond$}
\renewlist{itemize}{itemize}{5}

\algdef{SE}[PROCEDURE]{Procedure}{EndProcedure}%
[2]{\algorithmicprocedure\ \textproc{#1}\ifthenelse{\equal{#2}{}}{}{(#2)}}%
{\algorithmicend\ \algorithmicprocedure}%
   
\algdef{SE}[FUNCTION]{Function}{EndFunction}%
[2]{\algorithmicfunction\ \textproc{#1}\ifthenelse{\equal{#2}{}}{}{(#2)}}%
{\algorithmicend\ \algorithmicfunction}%

\usepackage{hyperref}
\hypersetup{
    colorlinks,
    citecolor=black,
    filecolor=black,
    linkcolor=blue,
    urlcolor=black}
    
\usepackage[export]{adjustbox}

\def\qed{\fbox{}}

\def\N{\mathcal{N}}   % we don't need integers, so both N and NN are gaussian

\def\R{{\mathbb R}}

\begin{document}
%\title{Rescaling the Variance of Gaussian Sampling for Better Design of Experiments}
%\titlerunning{Rescaling the Variance of Gaussian Sampling for Better DoEs}
%\title{Initialize smaller in high dimension}
%\title{Rescaling in high-dimension}
\title{Variance Reduction for Better Sampling in Continuous Domains}
% \title{Reducing Variance improves Random Sampling in Large Dimensions}
%Continuous Domains: Initialize Smaller in High Dimension}
%\title{Rescale small population in high dimension}
%\title{Rescaling a Small Population in High Dimension in Continous Domains: $\sigma=\sqrt{\log(\lambda)/d}$}
\renewcommand{\topfraction}{.95}
%\title{Rescaling your population in high dimension for initialization and one-shot optimization in continuous domains}
%\title{Rescaling your population in high dimension for black-box optimization in continuous domains}
%
\author{Laurent Meunier\inst{1,2} \and
Carola Doerr\inst{3}\and 
Jeremy Rapin\inst{1} \and 
Olivier Teytaud\inst{1} 
}
\authorrunning{L. Meunier, C. Doerr, J. Rapin, O.Teytaud}
\institute{Facebook Artificial Intelligence Research (FAIR), Paris, France
\and 
PSL, Université Paris-Dauphine, Miles Team
\and
Sorbonne Université, CNRS, LIP6, Paris, France\\
%\email{carola.doerr@lip6.fr}
}
%\author{Laurent Meunier}
%\author{Carola Doerr}
%\author{Jeremy Rapin}
%\author{Olivier Teytaud}
% LMeunier
% CDoerr
% OTeytaud
% JRapin
\let\oldmaketitle\maketitle
\renewcommand{\maketitle}{\oldmaketitle\setcounter{footnote}{0}}
\maketitle
\begin{abstract}
% Recent papers suggested that random search with $\lambda$ points should be rescaled closer to the center when the dimension becomes large compared to the logarithm of the budget $\lambda$. We theoretically and experimentally investigate this question. Theoretical results are applicable to fully parallel optimization (i.e. one single epoch for evolutionary computation, also known as design of experiments), and recommend to sample a distribution rescaled with ratio $\min(1,\Theta(\sqrt{\log(\lambda)/d}))$ - i.e. more peaked towards the center in high dimension. They are also applicable to the initialization of evolutionary algorithms.
Design of experiments, random search, initialization of population-based methods, or sampling inside an epoch of an evolutionary algorithm use a sample drawn according to some probability distribution for approximating the location of an optimum. Recent papers have shown that the optimal {\textit{search}} distribution, used for the sampling, might be more peaked around the center of the distribution than the {\textit{prior}} distribution modelling our uncertainty about the location of the optimum.
We confirm this statement, provide explicit values for this reshaping of the search distribution depending on the population size $\lambda$ and the dimension $d$, and validate our results experimentally.
\end{abstract}
%\setcounter{tocdepth}{5}
%\tableofcontents
\section{Introduction}
%\carolanote{Please read this very critically and feel free to edit. This intro is still far from being a Masterpiece}
We consider the setting in which one aims to locate an optimal solution $x^* \in \R^d$ for a given black-box problem $f:\R^d \to \R$ 
%that is known to follow the standard normal distribution $\N(0,I_d)$. 
through a parallel evaluation of $\lambda$ solution candidates. A simple, yet effective strategy for this \textit{one-shot optimization} setting is to choose the $\lambda$ candidates from a normal distribution $\N(\mu,\sigma^2)$, typically centered around an \textit{a priori} estimate $\mu$ of the optimum and using a variance $\sigma^2$ that is calibrated according to the uncertainty with respect to the optimum. 
%that the optimum is indeed close to $\mu$. 
\olivier{Random independent} sampling is -- despite its simplicity -- still a very commonly used and performing good technique in one-shot optimization settings. There also exist more sophisticated sampling strategies like Latin Hypercube Sampling (LHS~\cite{LHS}), or quasi-random constructions such as Sobol, Halton, Hammersley sequences~\cite{DickP10,Mat99} -- see~\cite{bergstra,icmldoe} for examples. However, no general superiority of these strategies over random sampling can be observed when the benchmark set is sufficiently diverse~\cite{BossekKNND19}. It is therefore not surprising that in several one-shot settings -- for example, the design of experiments~\cite{nie,mckay,hammersley,atanasov} or the initialization (and sometimes also further iterations) of evolution strategies -- the solution candidates are \olivier{frequently} sampled from \olivier{random independent} distributions \olivier{(though sometimes improved by mirrored sampling~\cite{antithetic})}. 
A surprising finding was recently communicated in~\cite{icmldoe}, where Cauwet et al. consider the setting in which the optimum $x^*$ is known to be distributed according to a standard normal distribution $\N(0,I_d)$, and the goal is to minimize the distance of the best of the $\lambda$ samples to this optimum. In the context of evolution strategies, one would formulate this problem as minimizing the sphere function with normally distributed optimum. Intuitively, one might guess that sampling the $\lambda$ candidates from the same prior distribution, $\mathcal{N}(0,I_d)$, should be optimal. This intuition, however, was disproved in~\cite{icmldoe}, where it is shown that -- unless the sample size $\lambda$ grows exponentially fast in the dimension $d$ -- the median quality of sampling from $\N(0,I_d)$ is worse than that of sampling a single point, \carola{namely} the center point $0$. A similar observation was previously made in~\cite{centerbased}, without \carola{mathematically proven guarantees.} %any proof.
\paragraph{Our Theoretical Result.} It was left open in~\cite{icmldoe} how to optimally scale the variance $\sigma^2$ when sampling the $\lambda$ solution candidates from a normal distribution $\N(0,\sigma^2 I_d)$. While the result from~\cite{icmldoe} suggests to use $\sigma=0$, we show in this work that a more effective strategy exists. More precisely, we show that setting $\sigma^2=\min\{1,\Theta(\log(\lambda)/d)\}$ is asymptotically optimal, as long as $\lambda$ is sub-exponential, but growing in $d$. Our variance scaling factor reduces the median approximation error by a $1-\epsilon$ factor, with $\epsilon=\Theta(\log(\lambda)/d)$. We also prove that no constant variance nor any other variance scaling as $\omega(\log(\lambda)/d)$ can achieve such an approximation error. Note that several optimization algorithms operate with rescaled sampling. Our theoretical results  therefore set the mathematical foundation for empirical rules of thumb \olivier{such as, for example, used in e.g.~\cite{centerbased,recenterbase,papapam,r1,r2,r3,icmldoe}. }
\paragraph{Our Empirical Results.} 
We complement our theoretical analyses by an empirical investigation of the rescaled sampling strategy. Experiments on the sphere function confirm \olivier{the results}. We also show that our scaling factor for the variance yields excellent performance on two other benchmark problems, the Cigar and the Rastrigin function. Finally, we demonstrate that these improvements are not restricted to the one-shot setting, but extend to iterative \carola{optimization} %sampling 
strategies. More precisely, we show a positive impact on the initialization of Bayesian optimization algorithms~\cite{ego} and on differential evolution~\cite{de}.   
\paragraph{Related Work.} While the most relevant works for our study have been mentioned above, we briefly note that a similar surprising effect as observed here is the ``Stein phenomenon''~\cite{Stein56,JamesStein61}. Although an intuitive way to estimate the mean of a standard gaussian distribution is to compute the empirical mean, Stein showed that this strategy is sub-optimal w.r.t. mean squared error and that the empirical mean needs to be rescaled by some factor to be optimal.
\section{Problem Statement and Related Work}\label{statement}
\label{sec:basics}
The context of our \otc{theoretical analysis} is \textit{one-shot optimization}. In one-shot optimization, we are allowed to select $\lambda$ points $x_1, \ldots, x_{\lambda} \in \R^d$. The quality $f(x_i)$ of these points is evaluated, and we measure the performance of our samples in terms of simple regret~\cite{bubeck2009pure} 
$\min_{i=1, \ldots,\lambda}f(x_i)-\inf_{x \in \R^d} f(x)$.\footnote{This requires knowledge of $\inf_{x} f(x)$, which may not be available in real-world \olivier{applications. In this case, the infimum can be replaced by an empirical minimum. In all} applications considered in this work the value of $\inf_x f(x)$ is known.} That is, we aim to minimize the distance -- measured in \textit{quality space} -- of the best of our points to the optimum. This formulation, however, also covers the case in which we aim to minimize the distance to the optimum in the \textit{search space}: we simply take as $f$ the root of the sphere function $f_{x^*}: \R^d \to \R, x \mapsto ||x - x^*||^2$, where here and in the following $||.||$ denotes the Euclidean norm. 
\paragraph{Rescaled Random Sampling for Randomly Placed Optimum.} In the setting studied in Sec.~\ref{sec:theory} we assume that the optimum $x^*$ is sampled from the standard multivariate Gaussian distribution $\N(0,I_d)$, and that we aim to minimize the \olivier{regret} $\min_{i=1,\ldots,\lambda} ||x_i-x^*||^2$ through i.i.d. samples $x_i \sim \N(0,\sigma^2 I_d)$. That is, in contrast to the classical \textit{design of experiments} (DoE) setting, we are only allowed to choose the scaling factor $\sigma$, whereas in DoE more sophisticated (often quasi-random and space-filling designs -- which are typically not i.i.d. samples\olivier{)} are admissible.
Intuitively, one might be tempted to guess that $\sigma=1$ should be a good choice, as in this case the $\lambda$ points are chosen from the same distribution as the optimum $x^*$. This intuition, however, was refuted in~\cite[Theorem~1]{icmldoe}, where is was shown that the middle point sampling strategy, which uses $\sigma=0$ (i.e., all $\lambda$ points collapse to $(0,\ldots,0)$) yields smaller regret than sampling from $\N(0,I_d)$ unless $\lambda$ grows exponentially in $d$. More precisely, it is shown in~\cite{icmldoe} that, for this regime of $\lambda$ and $d$, the median of $||x^*||^2$ is smaller than the median of $||x_i - x^*||^2$ for i.i.d. $x_i \in \N(0,I_d)$.  
This shows that sampling a single point can be better than sampling $\lambda$ points with the wrong scaling factor, unless the budget $\lambda$ is very large. \\
%Naturally, the regret of the middle point strategy is $||x^*||$, which follows the central $\chi^2$ distribution and has expectation $d$.
Our goal is to improve upon the middle point strategy, by deriving a scaling factor $\sigma$ such that the $\lambda$ i.i.d. samples yield smaller regret with a decent probability. More precisely, we aim at identifying $\sigma$ such that
\begin{equation}
\mathbb{P}\left[\min_{1\leq i \leq \lambda} ||x_i-x^*||^2\leq (1-\epsilon){||x^*||}^2\right]\geq \delta,
\label{beginning}
\end{equation}
for some $\delta \ge 1/2$ and $\epsilon >0$ as large as possible. \olivier{Here, }in line with~\cite{icmldoe}, we have switched \olivier{to regret}, for convenience of notation. \olivier{\cite{icmldoe} proposed, without proof, such a scaling factor: our proposal is dramatically better in some regimes.}
\section{Theoretical Results}
\label{sec:theory}
We derive sufficient and necessary conditions on the scaling factor $\sigma$ such that Eq.~\eqref{beginning} can be satisfied. More precisely, we prove that 
Eq.~\eqref{beginning} holds with approximation gain $\epsilon \approx \log(\lambda)/d$ when the \carola{variance} %rescaling factor 
$\sigma^2$ is chosen proportionally to $\log \lambda/d$ (and $\lambda$ does not grow too rapidly in $d$). We then show that Eq.~\eqref{beginning} cannot be satisfied for $\sigma^2 = \omega(\log(\lambda)/d)$. Moreover, we prove that $\epsilon = O(\log(\lambda)/d)$, which, together with the first result, shows that our scaling factor is asymptotically optimal. The precise statements are summarized in Theorems~\ref{thm:suff}, \ref{thm:nec}, and~\ref{thm:approx}, respectively. Proof sketches are available in Sec.~\ref{sketch}. Full proofs are left in the appendix.
\begin{theorem}{(Sufficient condition on rescaling)}
%\laurentnote{I changed a few things in the statement}
\label{thm:suff}
Let $\delta\in[\frac12,1)$. 
Let $\lambda=\lambda_d$, satisfying \inlineequation[ass3b]{\lambda_d\to\infty \text{ as } d\to\infty \text{ and } \log(\lambda_d)\in o(d)}.
% \begin{eqnarray}
% \lambda_d\to\infty \text{ as } d\to\infty \text{~~~~and~~~~}
% \log(\lambda_d)\in o(d).\label{ass3b}
% \end{eqnarray}
Then there exist two positive constants $c_1$, $c_2$, and $d_0$, such that for all $d\geq d_0$ it holds that \inlineequation[eq:betterdelta]{\mathbb{P}\left[\min_{i=1,\ldots,\lambda}||x^*-x_i||^2 \leq \left(1-\epsilon\right)||x^*||^2\right]\geq \delta}
% \begin{eqnarray}
% \mathbb{P}\left[\min_{i=1,\ldots,\lambda}||x^*-x_i||^2 \leq \left(1-\epsilon\right)||x^*||^2\right]\geq \delta,
% \label{eq:betterdelta}
% \end{eqnarray}
when $x^*$ is sampled from the standard Gaussian distribution $\mathcal{N}(0,I_d)$, 
$x_1,\ldots, x_{\lambda}$ are independently sampled from $\mathcal{N}(0,\sigma^2 I_d)$ with $\sigma^2 = \sigma^2_d=c_2 \log(\lambda)/d$ and 
$\epsilon=\epsilon_d=c_1 \log(\lambda)/d$. 
\end{theorem}
Theorem~\ref{thm:suff} shows that i.i.d. Gaussian sampling can outperform the middle-point strategy derived in~\cite{icmldoe} (i.e., the strategy using $\sigma^2=0$) if the scaling factor $\sigma$ is chosen appropriately. Our next theorem summarizes our findings for the conditions that are \textit{necessary} for the scaling factor $\sigma^2$ to outperform this middle-point strategy. This result, in particular, illustrates why neither the natural choice $\sigma=1$, nor any other constant scaling factor can be optimal.
% \begin{theorem}{(Necessary condition on rescaling)}
% \label{thm:nec}
% \olivier{For every $x^*$, let $f_{x^*}:x\in \mathbb{R}^d\rightarrow ||x-x^*||^2$, the square of the distance between $x$ and $x^*$. }We suppose that $x^*$ is distributed as a standard Gaussian distribution $\mathcal{N}(0,I)$.  Let $\lambda=\lambda_d\in \mathbb{N}$. \olivier{We also assume Assumptions~\ref{ass1b}-~\ref{ass3b} still holds. 
% Under these assumptions, let $\delta\in[1/2,1)$, and }let us assume that there exists $\epsilon>0$ \olivier{such that for $d$ sufficiently large:}
% \begin{equation}
%     P\left[\min_{i\in[\lambda_d]}||x^*-x_i||^2 \leq \left(1-\epsilon\right)||x^*||^2\big| x^*\right]\geq \delta
% \end{equation}
% with $x_i$ independently drawn from $N(0,\sigma_d^2 I)$. Then, \olivier{
% \begin{equation}\sigma^2=\sigma_d^2 \in O\left( \frac{\log\lambda_d}{d}\right).\label{eq6bis}\end{equation}}
% \end{theorem}
\begin{theorem}{(Necessary condition on rescaling)}
\label{thm:nec}
Consider $\lambda=\lambda_d$ satisfying assumptions~\eqref{ass3b}. %and  
There exists an absolute constant $C>0$ such that for all $\delta\in[\frac12,1)$, there exists $d_0>0$ such that, for all $d>d_0$ and for all $\sigma$ the property \inlineequation[eq6bis]{\exists \epsilon>0,\mathbb{P}\left[\min_{i=1,\ldots,\lambda}||x^*-x_i||^2 \leq \left(1-\epsilon\right)||x^*||^2\right]\geq \delta}
% depending on $d$ and $\lambda$:  %\ref{eq:betterdelta}:
% \begin{small}
% \begin{equation}
% \exists \epsilon>0,\mathbb{P}\left[\min_{i=1,\ldots,\lambda}||x^*-x_i||^2\leq \left(1-\epsilon\right)||x^*||^2\right]\geq \delta,
% %\label{eq:betterdelta}
% %\end{eqnarray}
% %implies
% %\begin{equation}
% %log\lambda/d\in O\left( \frac{\log\lambda}{d}\right).
% \label{eq6bis}
% \end{equation}
% \end{small}
for $x^* \sim \N(0,I_d)$ and $x_1,\ldots, x_{\lambda}$ independently sampled from $\N(0,\sigma^2 I_d)$,
implies that $\sigma^2 \leq C \log(\lambda)/d$.
\end{theorem}
While Theorem~\ref{thm:nec} induces a necessary condition on the scaling factor $\sigma$ to improve over the middle-point strategy, it does not bound the gain that one can achieve through a proper scaling. Our next theorem shows that the factor derived in Theorem~\ref{thm:suff} is asymptotically optimal. 
\begin{theorem}{(Upper bound for the approximation factor)} Consider $\lambda=\lambda_d$ satisfying assumptions~\eqref{ass3b}. %and  
There exists an absolute constant $C'>0$ such that for all $\delta\in[\frac12,1)$, there exists $d_0>0$ such that, for all $d>d_0$ and for all $\epsilon,\sigma>0$, it holds that if $\mathbb{P}\left[\min_{i=1,\ldots,\lambda}||x^*-x_i||^2\leq \left(1-\epsilon\right)||x^*||^2\right]\geq \delta$
for $x^* \sim \N(0,I_d)$ and $x_1,\ldots, x_{\lambda}$ independently sampled from $\N(0,\sigma^2 I_d)$,
then $\epsilon\leq C' \log(\lambda)/d$.
\label{thm:approx}
\end{theorem}
\subsection{Proof Sketches}
\label{sketch}
% The full proofs of Theorems~\ref{thm:suff},~\ref{thm:nec} and~\ref{thm:approx} are in Appendices ADD REF with all technical details. We also remind in Appendix properties of $\chi^2$ distributions which are essential to our proofs. Here we add a fast skecth of the proofs. We just give the main ideas of the proofs, all technical  details are left in Appendix.\\
We first notice that as $x^*$ is sampled from a standard normal distribution $\mathcal{N}(0,I_d)$, its norm satisfies $||x^*||^2  = d + o(d)$ as $d\rightarrow\infty$. 
We then use that, conditionally to $x^*$, it holds that 
\begin{equation*}
\resizebox{.99\hsize}{!}{ $\mathbb{P}\left[\min_{i\in[\lambda]}||x^*-x_i||^2 \leq \left(1-\epsilon\right)||x^*||^2\big| x^*\right]  
=1-\left(1-\mathbb{P}\left[||x-x^*||^2 \leq \left(1-\epsilon\right)||x^*||^2\big|x^*\right]\right)^\lambda$}
\end{equation*}
% \begin{align*}
%  \mathbb{P}&\left[\min_{i\in[\lambda]}||x^*-x_i||^2 \leq \left(1-\epsilon\right)||x^*||^2\big| x^*\right] \\ 
% &=1-\left(1-\mathbb{P}\left[||x-x^*||^2 \leq \left(1-\epsilon\right)||x^*||^2\big|x^*\right]\right)^\lambda\,.  
% \end{align*}
We therefore investigate when the condition 
\begin{align}
\mathbb{P}\left[||x-x^*||^2 \leq \left(1-\epsilon\right)||x^*||^2\big|x^*\right]>1-(1-\delta)^{\frac{1}{\lambda}}
\label{eq:requiredCondi}
\end{align}
 is satisfied. 
To this end, we make use of the fact that  the squared distance $||x^*||^2$ of $x^*$ to the middle point $0$ follows the central $\chi^2(d)$ distribution, whereas, for a given point $x^* \in \R^d$, the distribution of the squared distance $||x - x^*||^2/\sigma^2$ for $x \sim \N(0,\sigma^2I_d)$ follows the non-central $\chi^2(d,\mu)$ distribution with non-centrality parameter $\mu:=||x^*||^2/\sigma^2$. Using the concentration inequalities provided in~\cite[Theorem 7]{chi2magical} for non-central $\chi^2$ distributions, we then derive sufficient and necessary conditions for condition~\eqref{eq:requiredCondi} to hold. 
With this, and using assumptions~\eqref{ass3b}, we are able to derive the results from Theorems~\ref{thm:suff}, \ref{thm:nec}, and~\ref{thm:approx}. 
\section{Experimental Performance Comparisons}
\label{sec:experiments}
\begin{figure}[t]
    \centering
    \begin{minipage}{0.72\textwidth}
        \centering
        \includegraphics[width=0.8\textwidth]{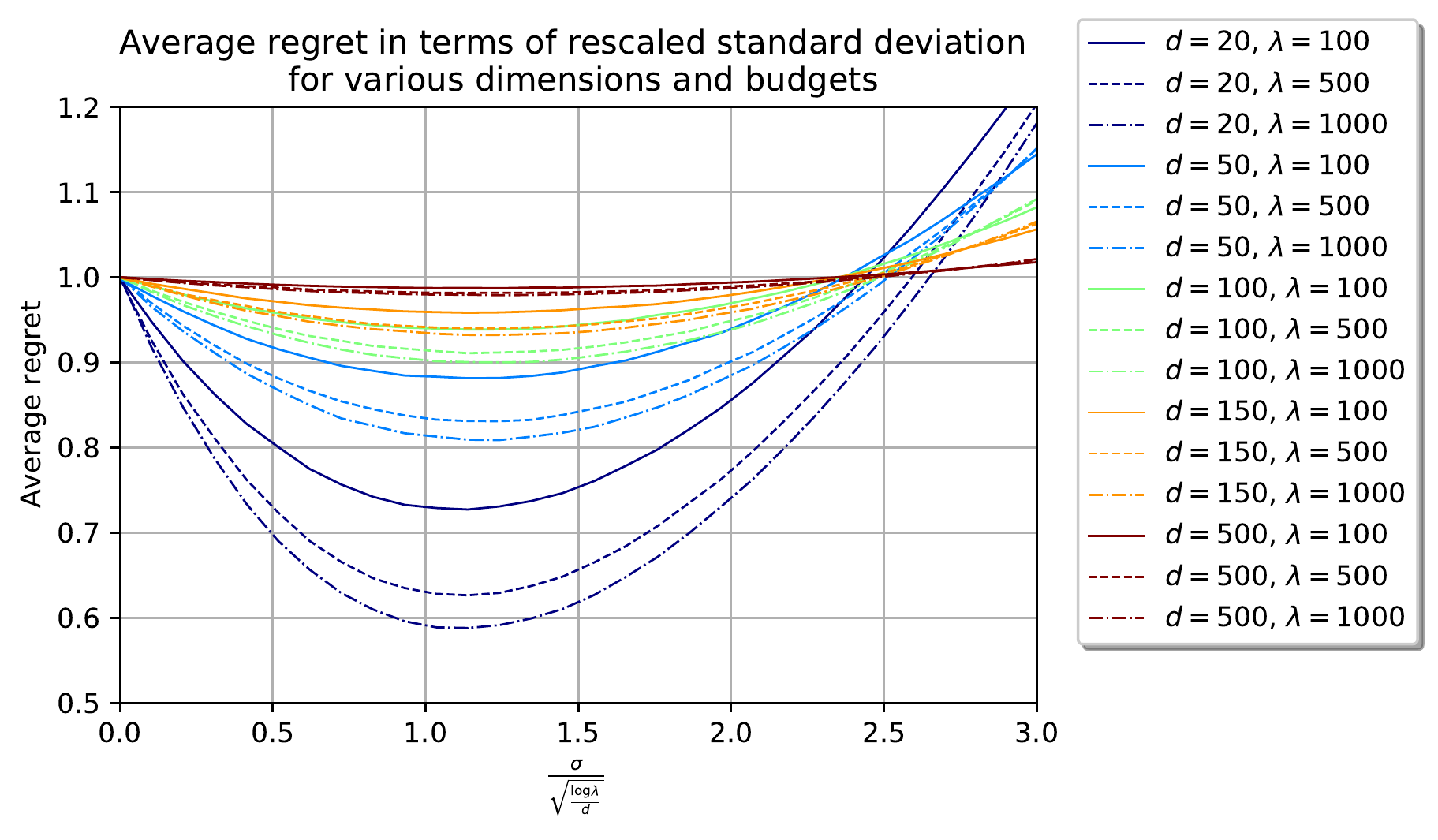} % first figure itself
        \caption{Average regret, normalized by $d$, on the sphere function for various dimensions and budgets in terms of rescaled standard deviation. Each mean has been estimated from $100,000$ samples. Table on the right: Average regret for $\sigma^*=\sqrt{\log(\lambda)/d}$ and $\sigma=1$.}
        \label{valsphere}
    \end{minipage}\hfill
    \begin{minipage}{0.25\textwidth}
           {\footnotesize{
        \centering
        % \begin{table}
    % \centering
    \begin{tabular}{|c|c||c|c|}
\hline
$d$&	 $\lambda$&  $\sigma^*$&    $\sigma=1$\\
\hline
\multirow{3}{*}{20}&     $100$&		$\mathbf{0.73}$&	$0.88$\\
&	    $500$&		$\mathbf{0.63}$&	$0.72$\\
&	    $1000$&		$\mathbf{0.59}$&	$0.66$\\
\hline
\multirow{3}{*}{50}&	    $100$&		$\mathbf{0.89}$&	$1.23$\\
&	    $500$&		$\mathbf{0.83}$&	$1.10$\\
& 	$1000$&		$\mathbf{0.81}$&	$1.05$\\
\hline
\multirow{3}{*}{100}&	$100$&		$\mathbf{0.94}$&	$1.44$\\
&    $500$&		$\mathbf{0.91}$&	$1.33$\\
&    $1000$&		$\mathbf{0.90}$&	$1.29$\\
\hline
\multirow{3}{*}{150}&	$100$&		$\mathbf{0.96}$&	$1.53$\\
&	$500$&		$\mathbf{0.94}$&	$1.44$\\
&	$1000$&		$\mathbf{0.93}$&	$1.41$\\
\hline
\multirow{3}{*}{500} &	$100$&    	$\mathbf{0.99}$&	$1.74$\\
&	$500$&    	$\mathbf{0.98}$&	$1.68$\\
&	$1000$&		$\mathbf{0.98}$&	$1.66$\\
\hline    
    \end{tabular} 
}}
    \end{minipage}
\end{figure}
\carola{The theoretical results presented above are in asymptotic terms, and do not specify the constants. We therefore complement our mathematical investigation with an empirical analysis of the rescaling factor. Whereas results for the setting studied in Sec.~\ref{sec:theory} are presented in Sec.~\ref{laurentsigma}, we show in Sec.~\ref{rg} that the advantage of our rescaling factor is not limited to minimizing the distance in search space. More precisely, we show that the rescaled sampling achieves good results also in a classical DoE task, in which we aim for minimizing the regret for the Cigar and for the Rastrigin functions. 
Finally, we investigate in Sec.~\ref{sec:initialization} the impact of initializing two common optimization heuristics, Bayesian Optimization (BO) and differential evolution (DE), by a population sampled from the Gaussian distribution $\N(0,\sigma^2 I_d)$ using our rescaling factor $\sigma={\sqrt{\log(\lambda)/ d}}$}.
% \carolanote{we should briefly say what we see. Also, check if we do Gaussian or quasi-random sampling in the experiments} 
\subsection{Validation of Our Theoretical Results on the Sphere Function}
\label{laurentsigma}
\begin{figure}[t]
    \centering
    \includegraphics[trim=5 5 5 5,clip,width=.5\textwidth]{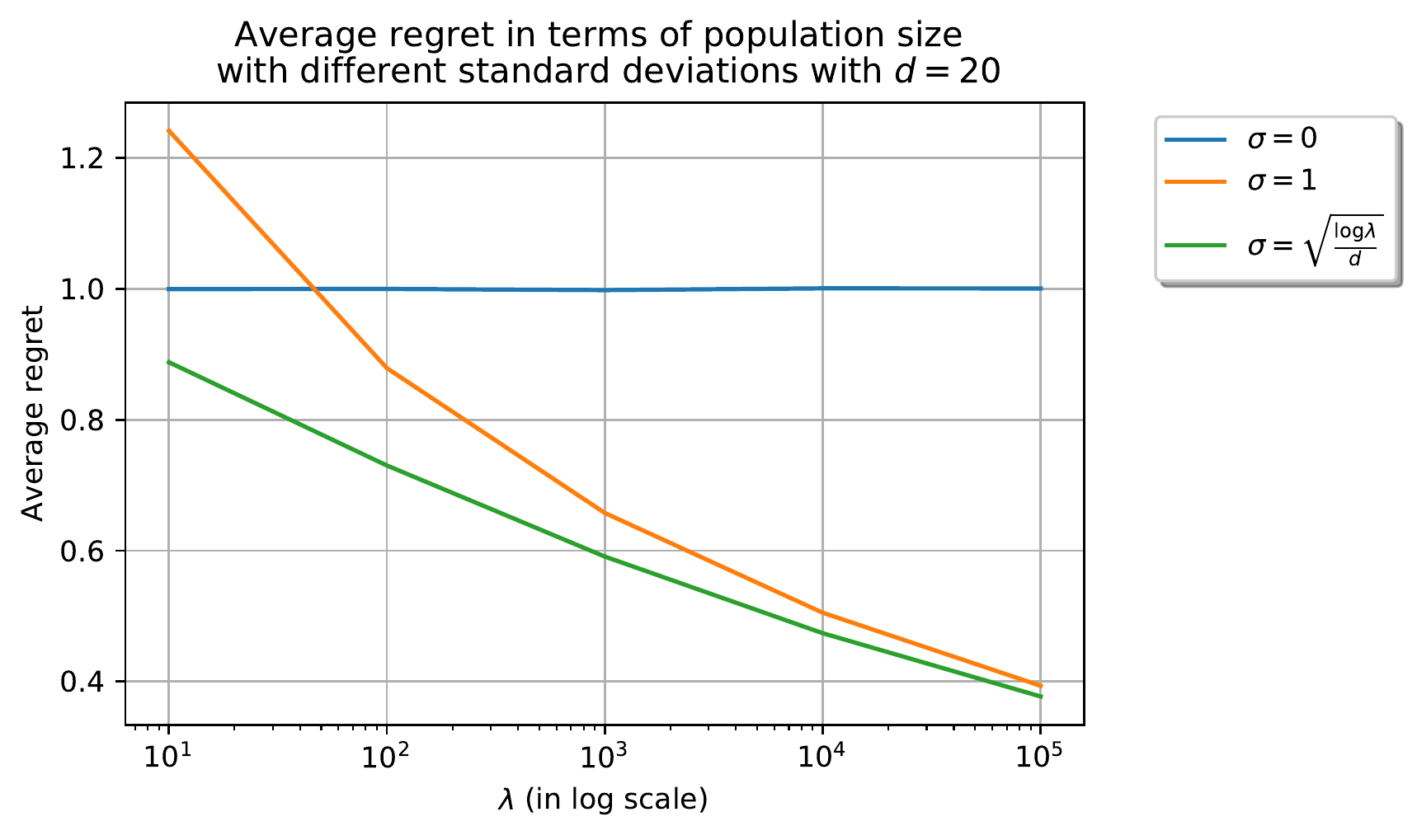}
    \includegraphics[width=.43\textwidth]{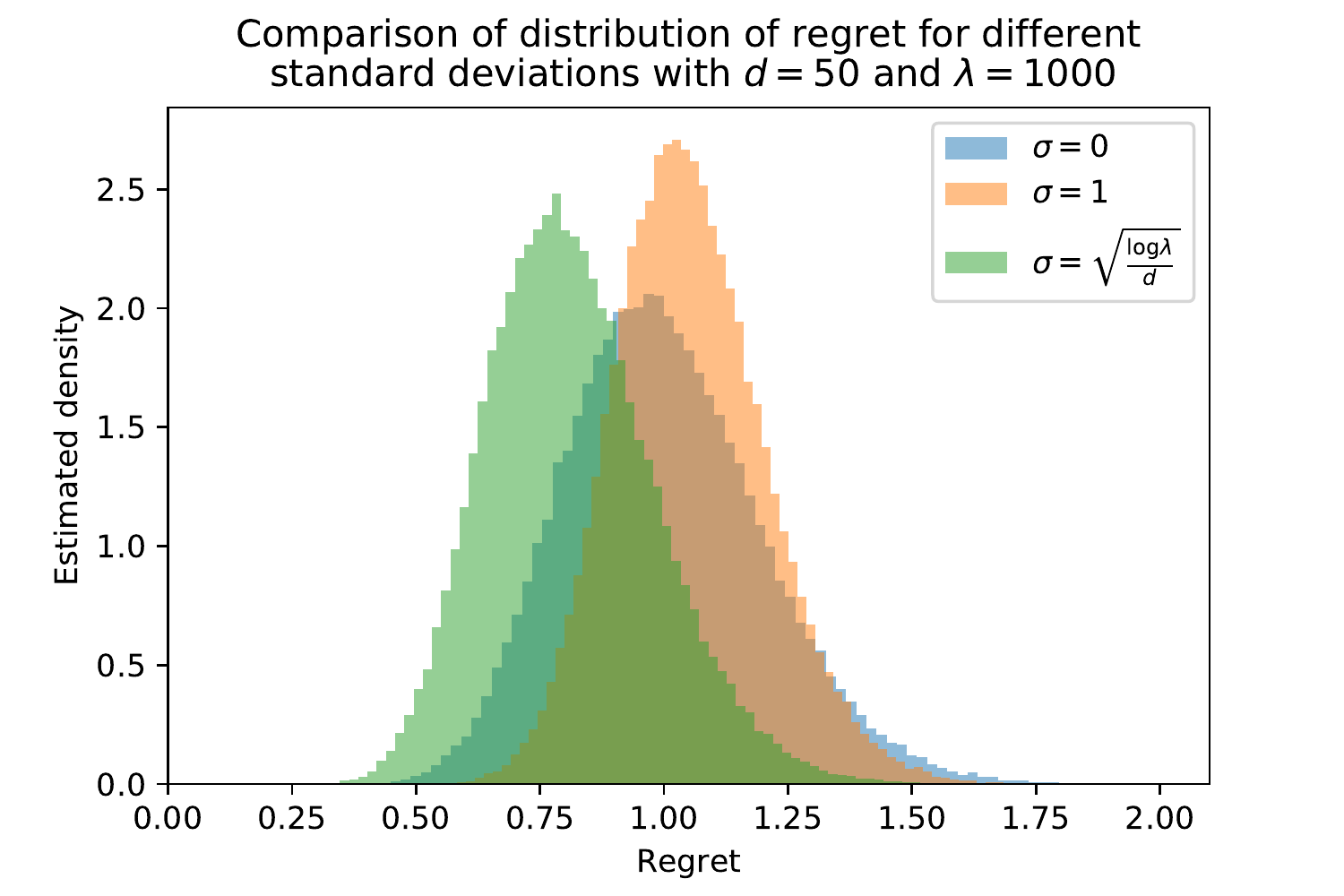}   \\ 
\caption{Comparison of methods: without rescaling ($\sigma=1$), midpoint sampling ($\sigma = 0$), and our rescaling method ($\sigma=\sqrt{\frac{\log\lambda}{d}}$). Each mean has been estimated from $10^5$ samples. (On left) Average regret, normalized by $d$, on the sphere function for diverse population sizes $\lambda$ at fixed dimension $d=20$. The gain of rescaling decreases as $\lambda$ increases.  (On right) Distribution of the regret for the strategies on the $50d$-sphere function for $\lambda=1000$. %\carola{We see that the dispersion of regret is similar.} %$\lambda\in\{500,1000\}$. \otc{We notice the variances are close.}
}
\label{fig:boxplots}
\end{figure}

%In this section, we lead experiments to validate our claim on the sphere function. 
Fig.~\ref{valsphere} displays the normalized \olivier{average regret} $\frac1d\mathbb{E}\left[\min_{i=1,\ldots,\lambda}||x^*-x_i||^2 \right] $ in terms of %$\frac{\sigma}{\sqrt{\log(\lambda)/d}}$ 
$\sigma/\sqrt{\log(\lambda)/d}$ 
for different dimensions and budgets. We observe that the best parametrization of $\sigma$ is around $\sqrt{\log(\lambda)/d}$ in all displayed cases. Moreover, we also see that -- as expected -- the gain of the rescaled sampling over the midpoint sampling ($\sigma=0$) goes to $0$ as $d\rightarrow\infty$. We also see that, for the regimes plotted in Fig.~\ref{valsphere}, the advantage of the rescaled variance grows with the budget~$\lambda$. Figure~\ref{fig:boxplots} (on left) displays the average regret as a function of increasing values of~$\lambda$ for the different rescaling methods ($\sigma\in\{0,\sqrt{\log \lambda/d},1\}$). We remark, unsurprisingly, that the gain of rescaling is diminishing as $\lambda\rightarrow\infty$. Finally, Figure~\ref{fig:boxplots} (on right) shows the distribution of regrets for the different rescaling methods. The improvement of the expected regret is not at the expense of a higher \carola{dispersion} %variance 
of the regret. 
\subsection{Comparison with the DoEs Available in Nevergrad}%Rescaled Sampling for DoE Tasks (aka One-Shot Optimization)}
\label{rg}
\begin{figure}[t]
    \centering
    \includegraphics[trim={10 5 12 80}, clip,width=0.75\textwidth]{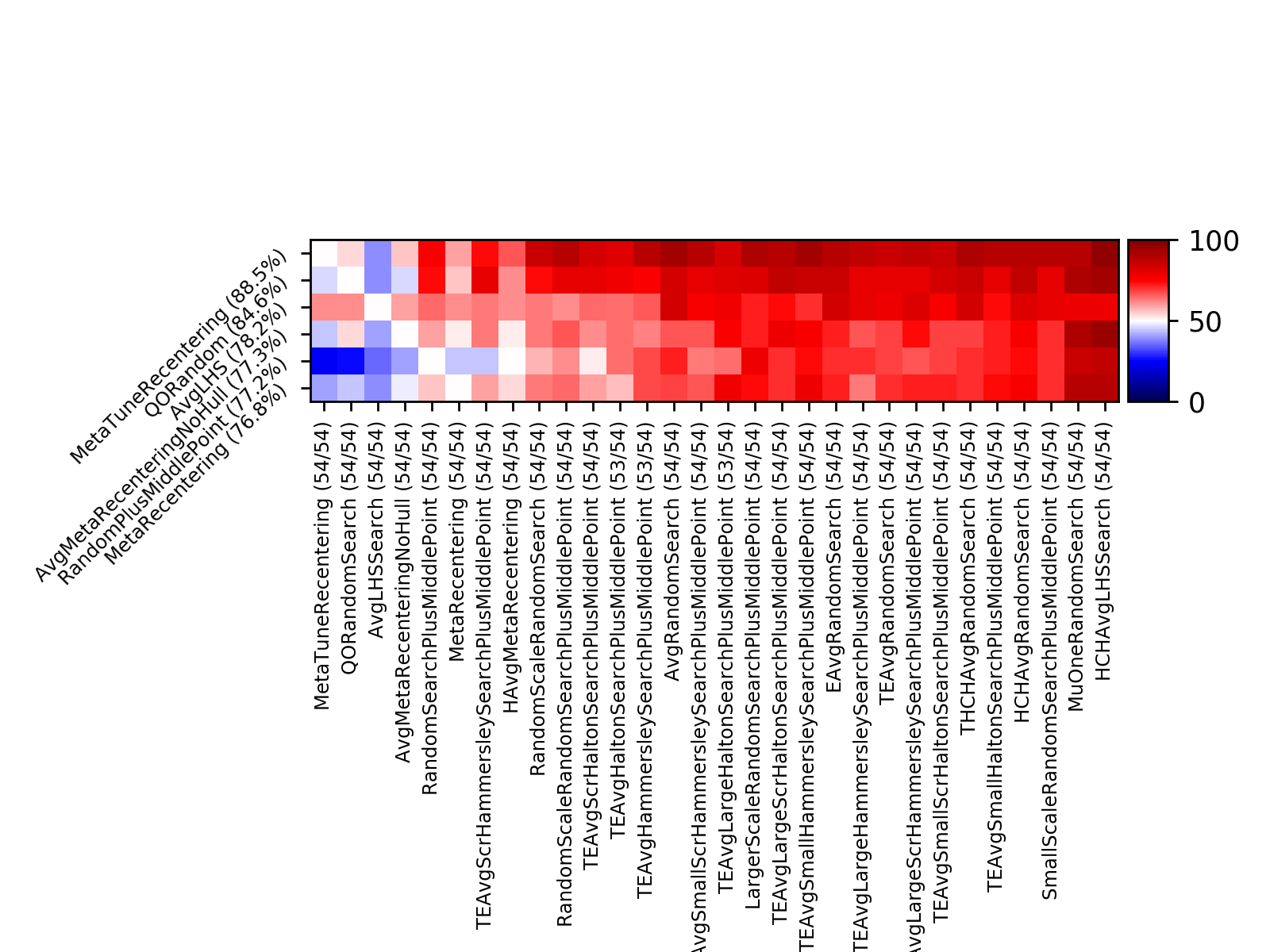}
    \caption{Comparison of various one-shot optimization methods from the point of view of the simple regret. Reading guide in Sec.~\ref{rg}. Results are averaged over objective functions Cigar, Rastrigin, Sphere in dimension $20$, $200$, $2000$, and budget $30$, $100$, $3000$, $10000$, $30000$, $100000$. \otc{\texttt{MetaTuneRecentering} performs best overall. Only the 30 best performing methods are displayed.}}
    \label{doexp}
\end{figure}
\begin{figure}[t]
    \centering
\includegraphics[trim={10 10 12 80}, clip,width=.32\textwidth]{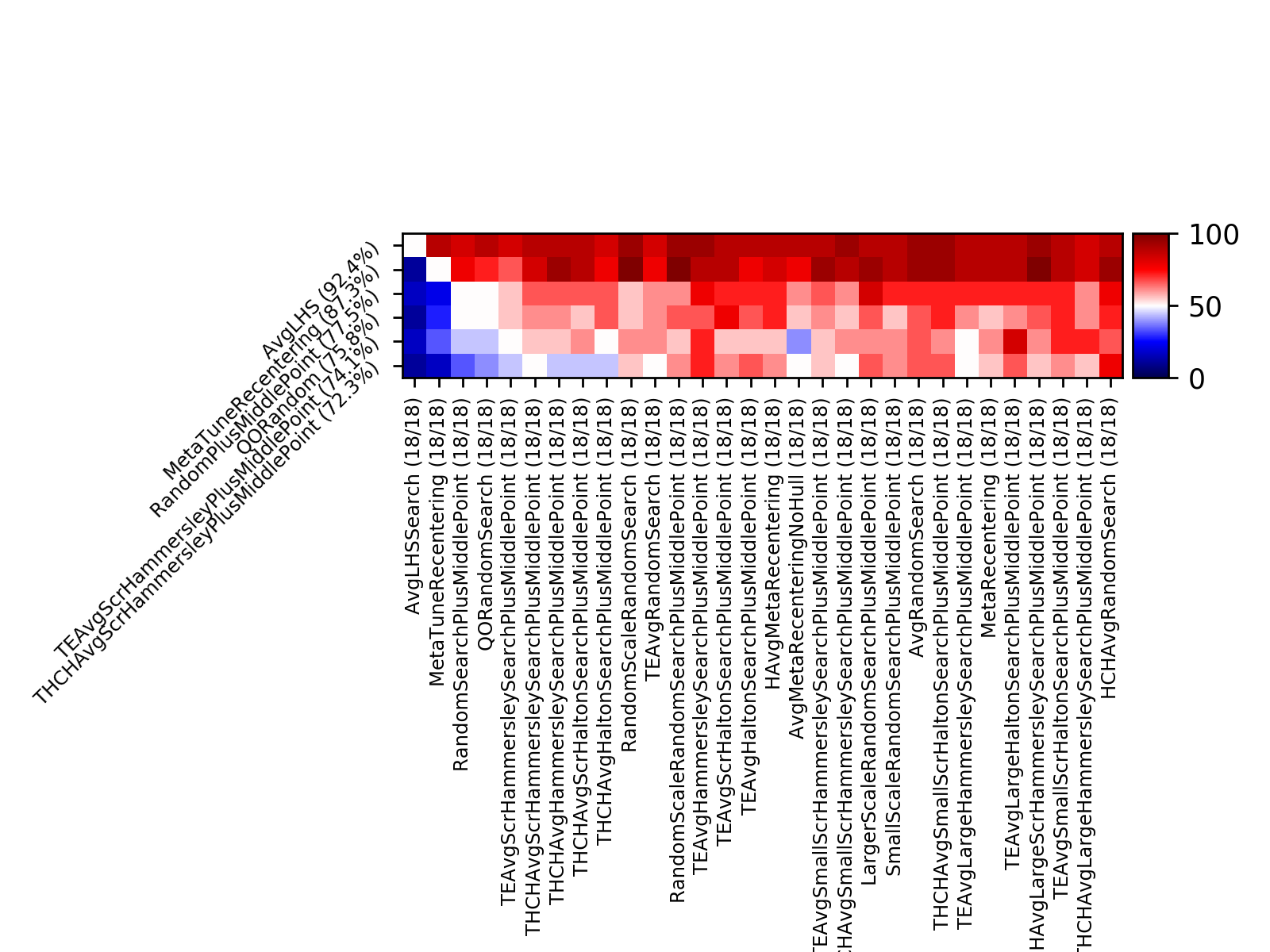}      \includegraphics[width=.32\textwidth]{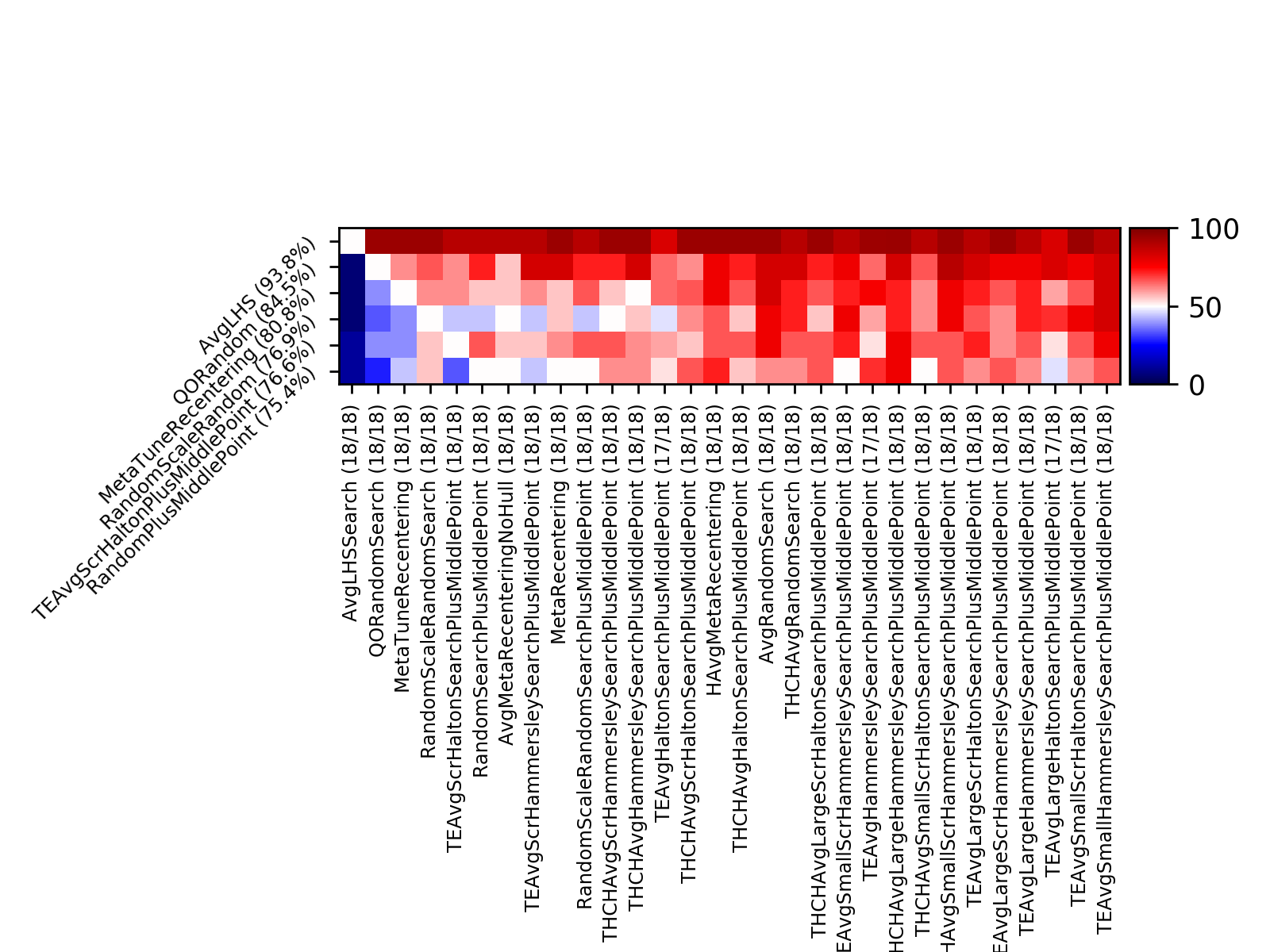}     
\includegraphics[trim={10 10 12 80}, clip,width=.32\textwidth]{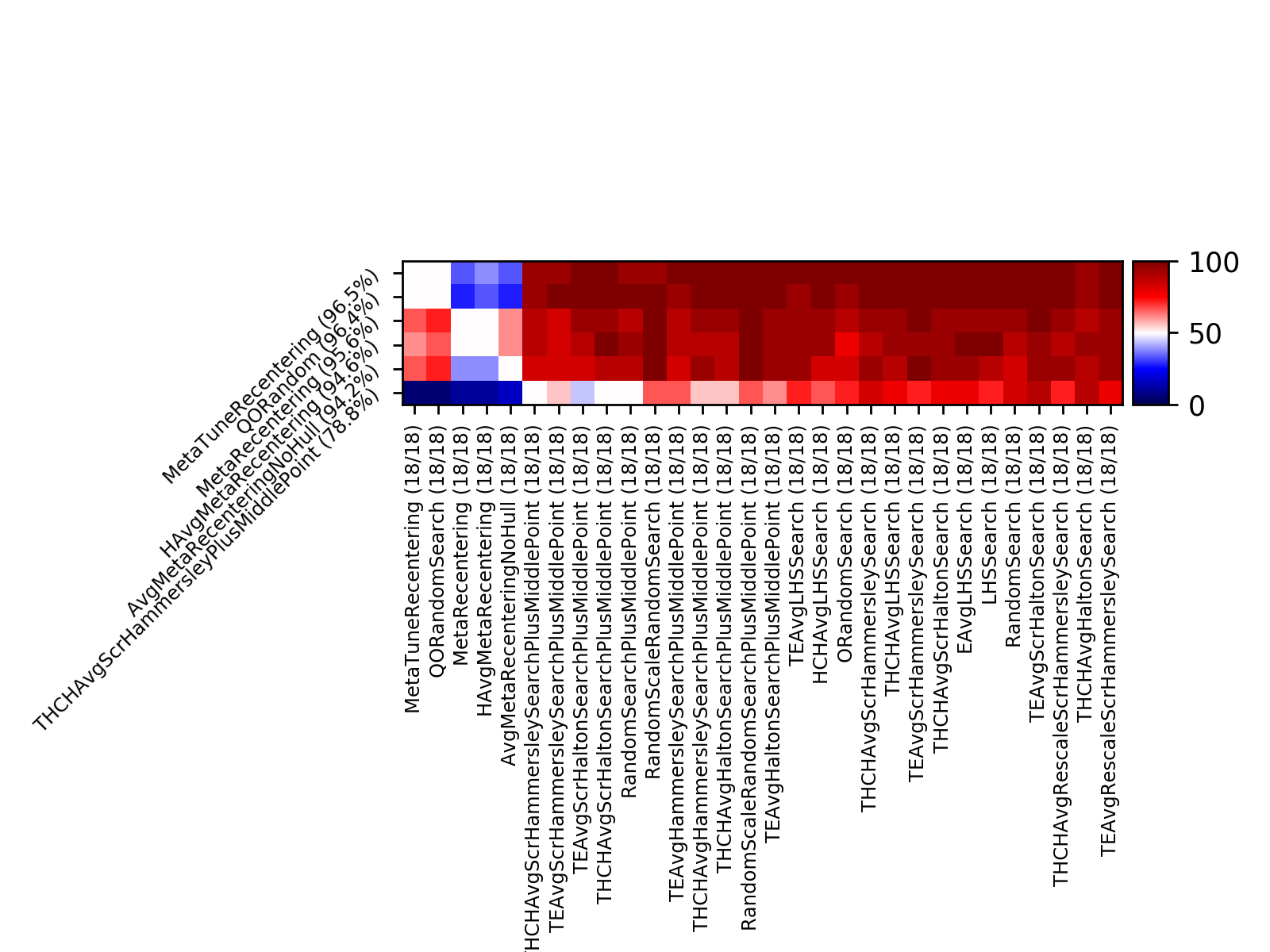}   \\ 
~\hfill Sphere function\hfill ~ \hfill Cigar function\hfill ~ \hfill Rastrigin function\hfill ~\\
\caption{Same experiment as Fig.~\ref{doexp}, but separately over each objective function. Results are still averaged over 6 distinct budgets ($30$, $100$, $3000$, $10000$, $30000$, $100000$) and 3 distinct dimensionalities ($20$, $200$, $2000$). \otc{\texttt{MetaTuneRecentering} performs well in each case, and is not limited to the sphere function for which it was derived. Variants of LHS are sometimes excellent and sometimes not visible at all (only the 30 best performing methods are shown).}}
    \label{toto3}
\end{figure}
\begin{figure}[t]
    \centering %top trimming was 97, but that removed too much
\includegraphics[trim={10 10 12 80}, clip,width=.48\textwidth]{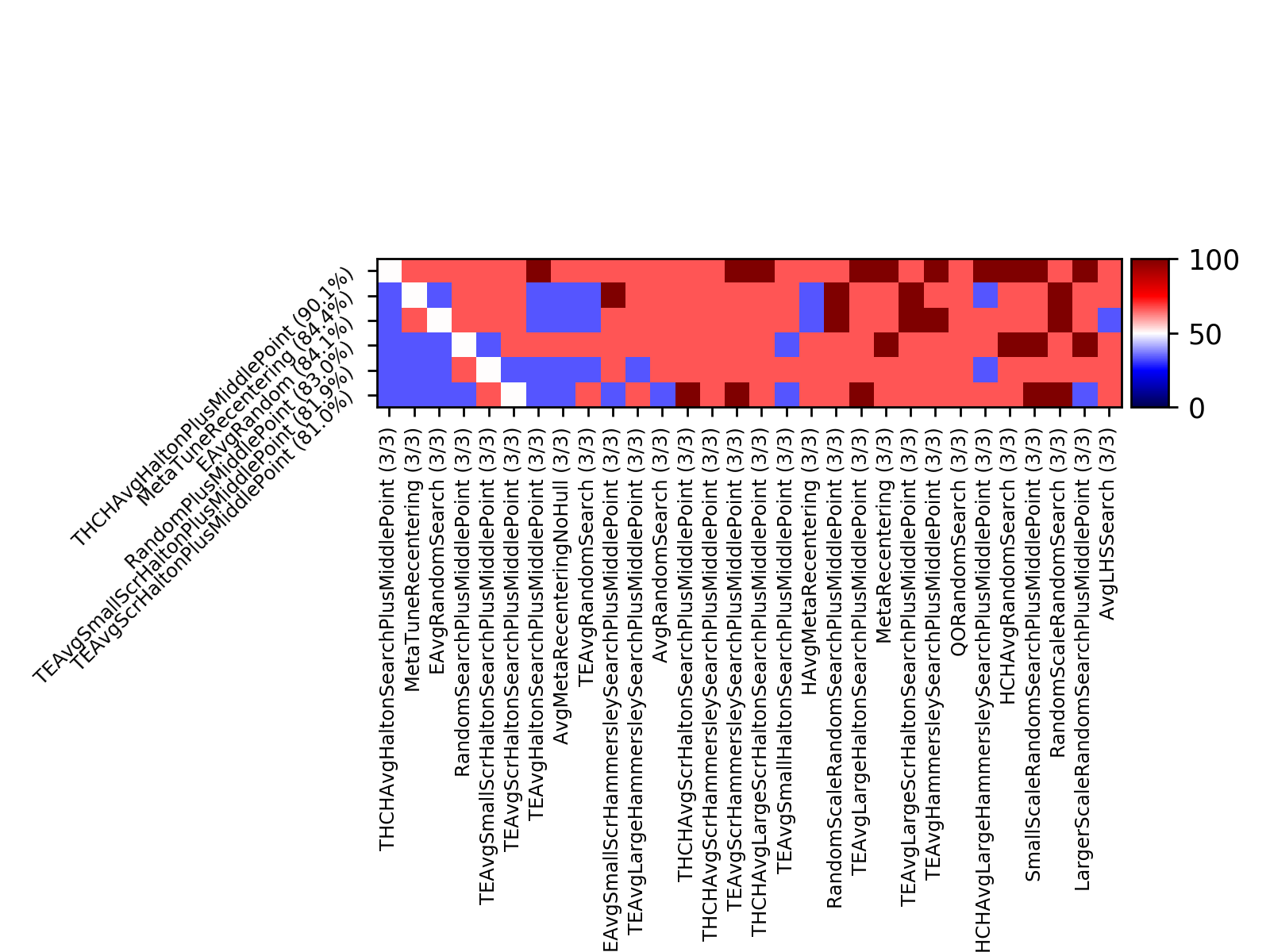}
\includegraphics[trim={10 10 12 80}, clip,width=.48\textwidth]{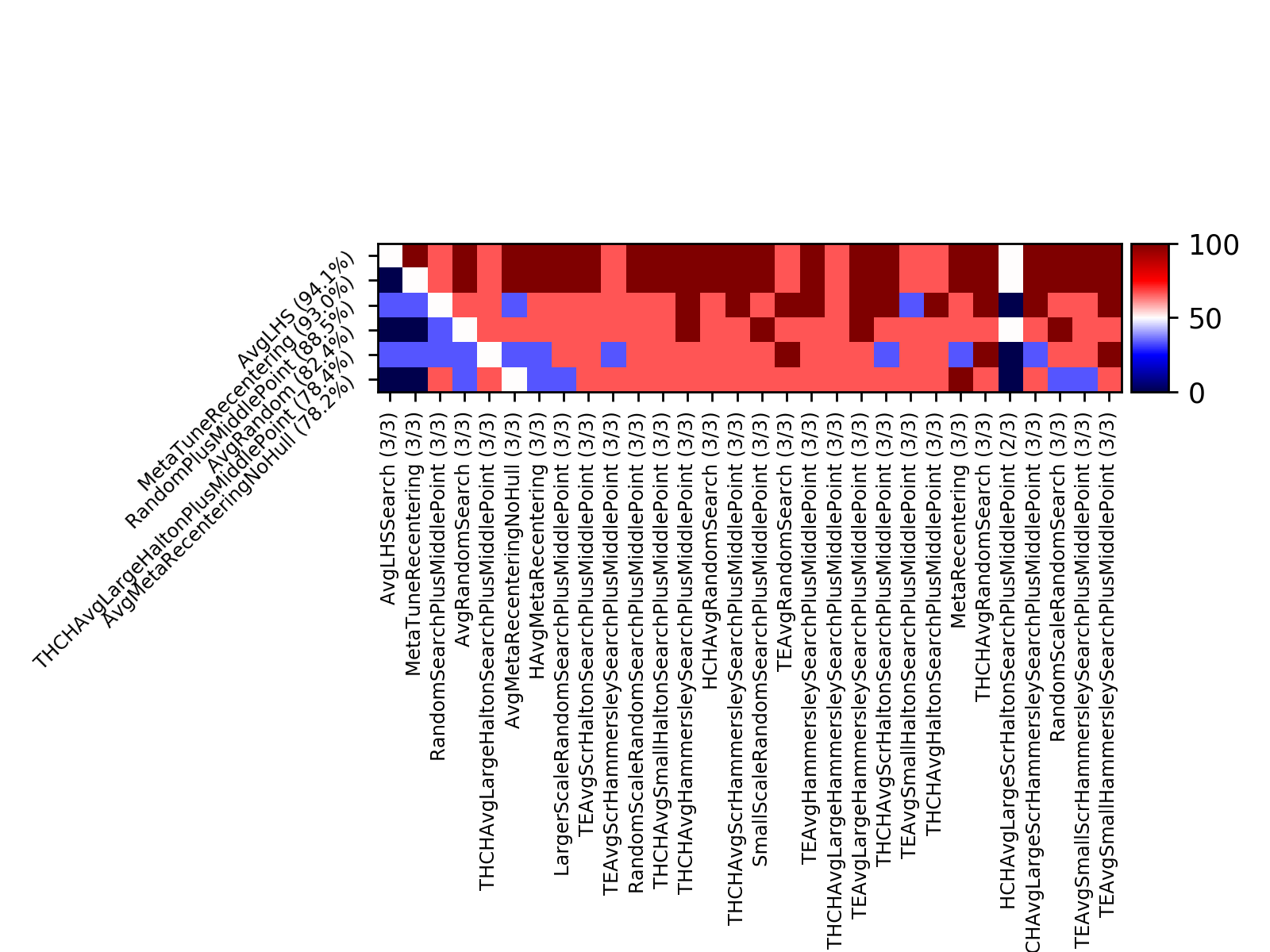}\\
~\hfill Budget $\lambda=30$ \hfill ~ \hfill Budget $\lambda=100$ \hfill ~ \\
\includegraphics[trim={10 10 12 80}, clip,width=.48\textwidth]{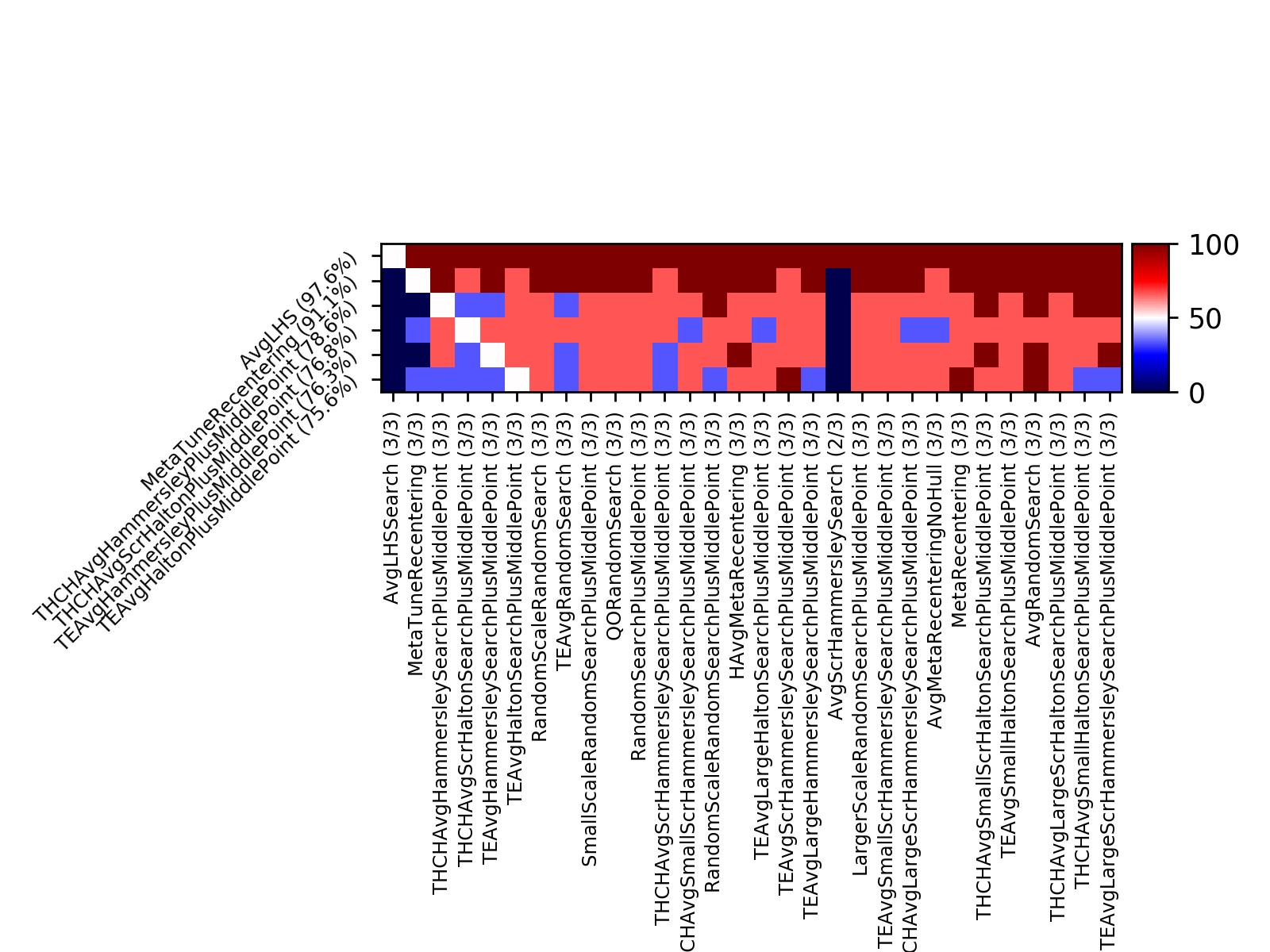}
\includegraphics[trim={10 10 12 80}, clip,width=.48\textwidth]{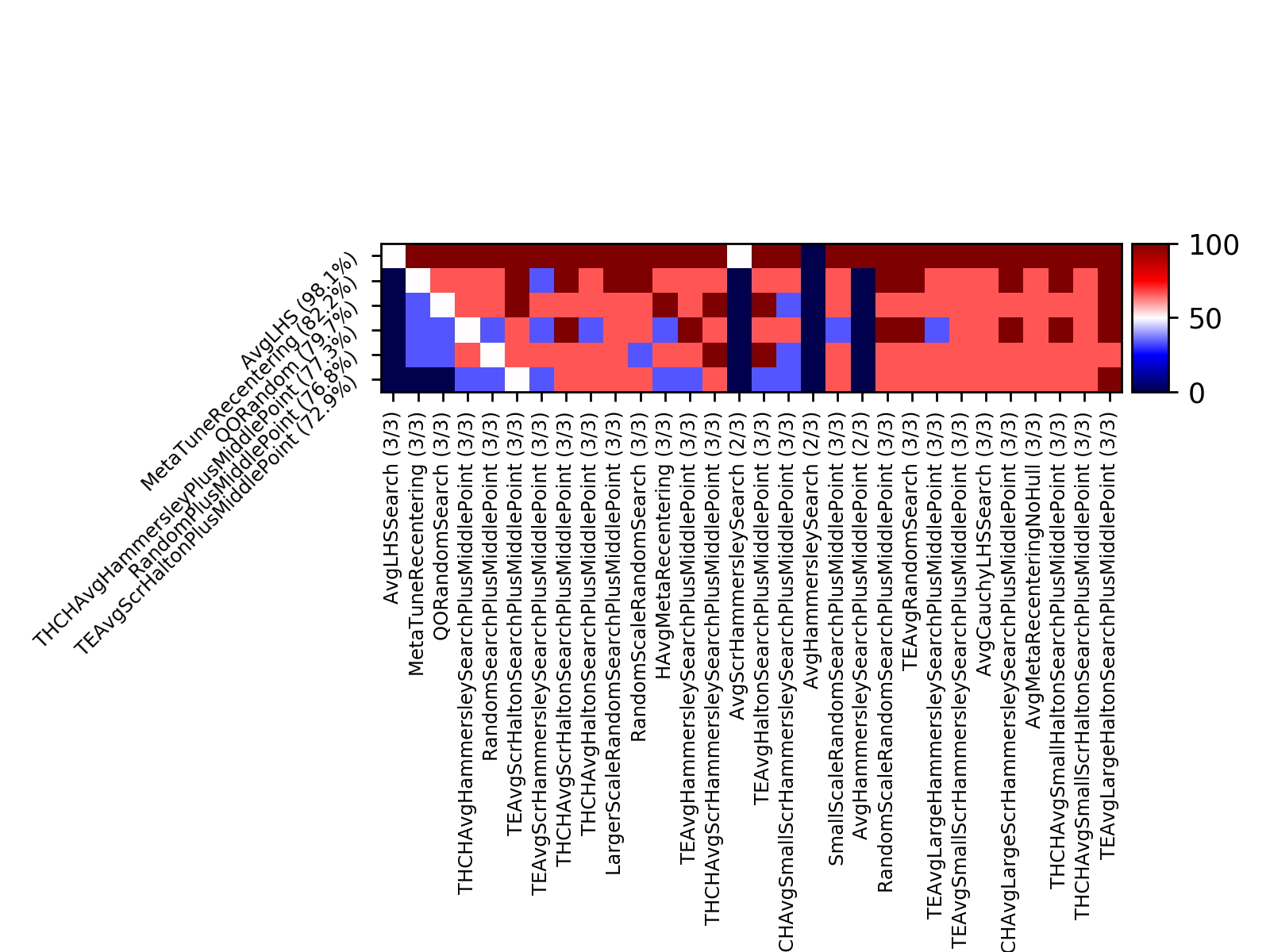}\\
~\hfill Budget $\lambda=3000$ \hfill ~ \hfill Budget $\lambda=10000$ \hfill ~ \\
\includegraphics[trim={10 10 12 80}, clip,width=.48\textwidth]{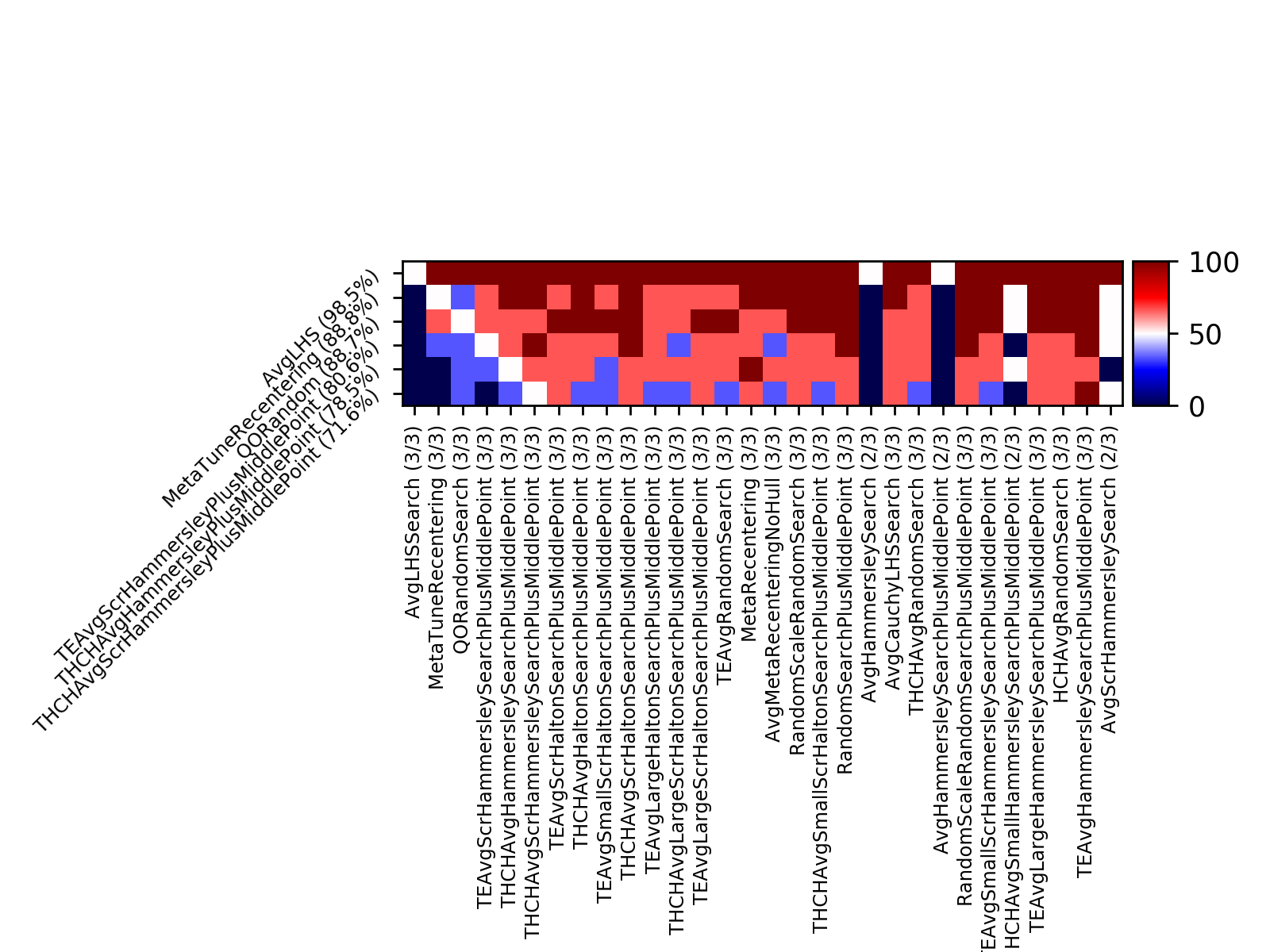}
\includegraphics[trim={10 10 12 80}, clip,width=.48\textwidth]{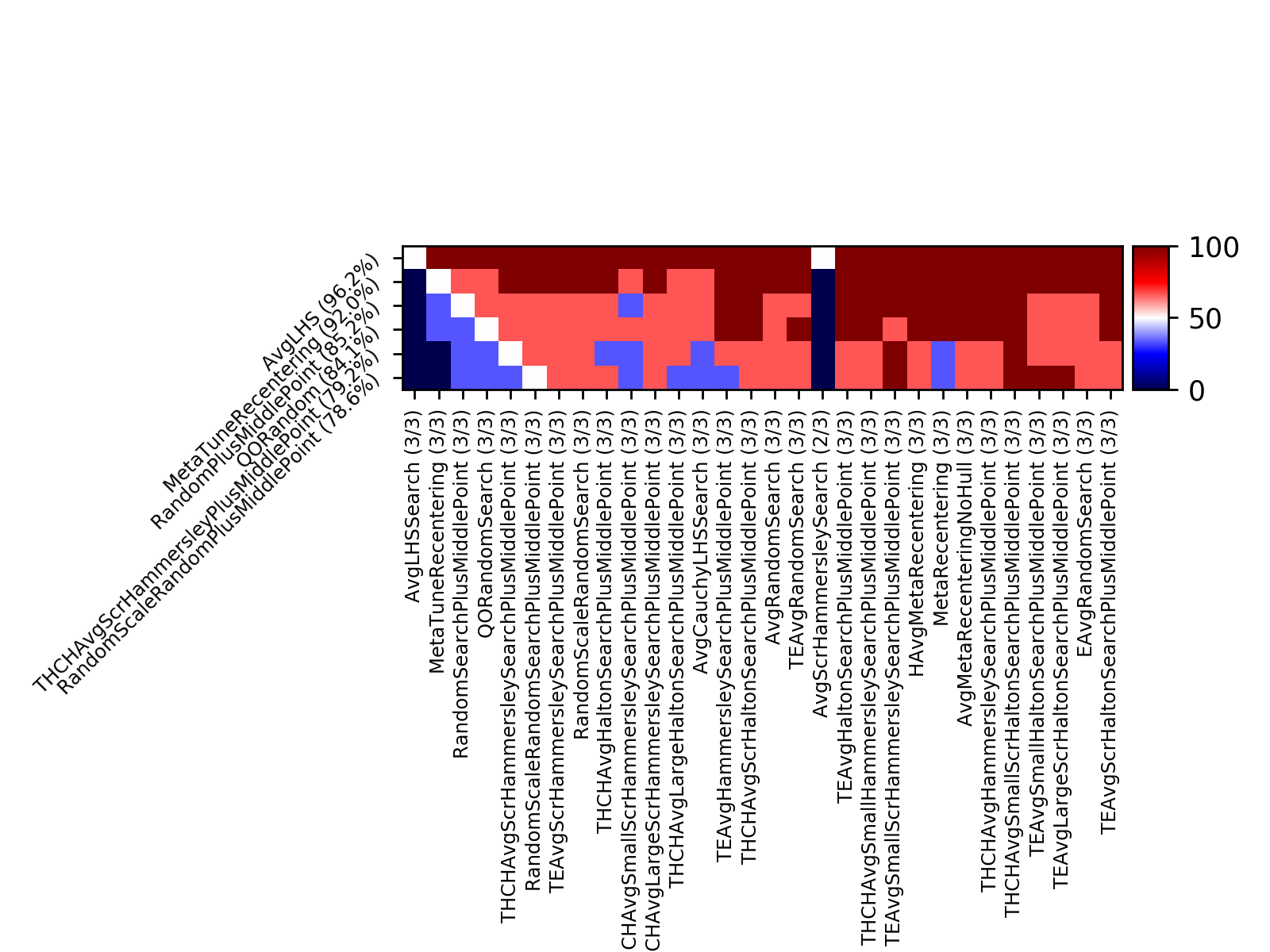}\\
~\hfill Budget $\lambda=30000$ \hfill ~ \hfill Budget $\lambda=100000$ \hfill ~ \\
\caption{Results on the sphere function, per budget. Results are averaged over dimension $20$, $200$, $2000$. Our method \texttt{MetaTuneRecentering} performs among the best in all cases. \otc{LHS is excellent on this very simple setting, namely the sphere function.}}
    \label{toto}
\end{figure}
\begin{figure}[t]
\centering
\includegraphics[trim={10 10 12 80}, clip,width=.32\textwidth]{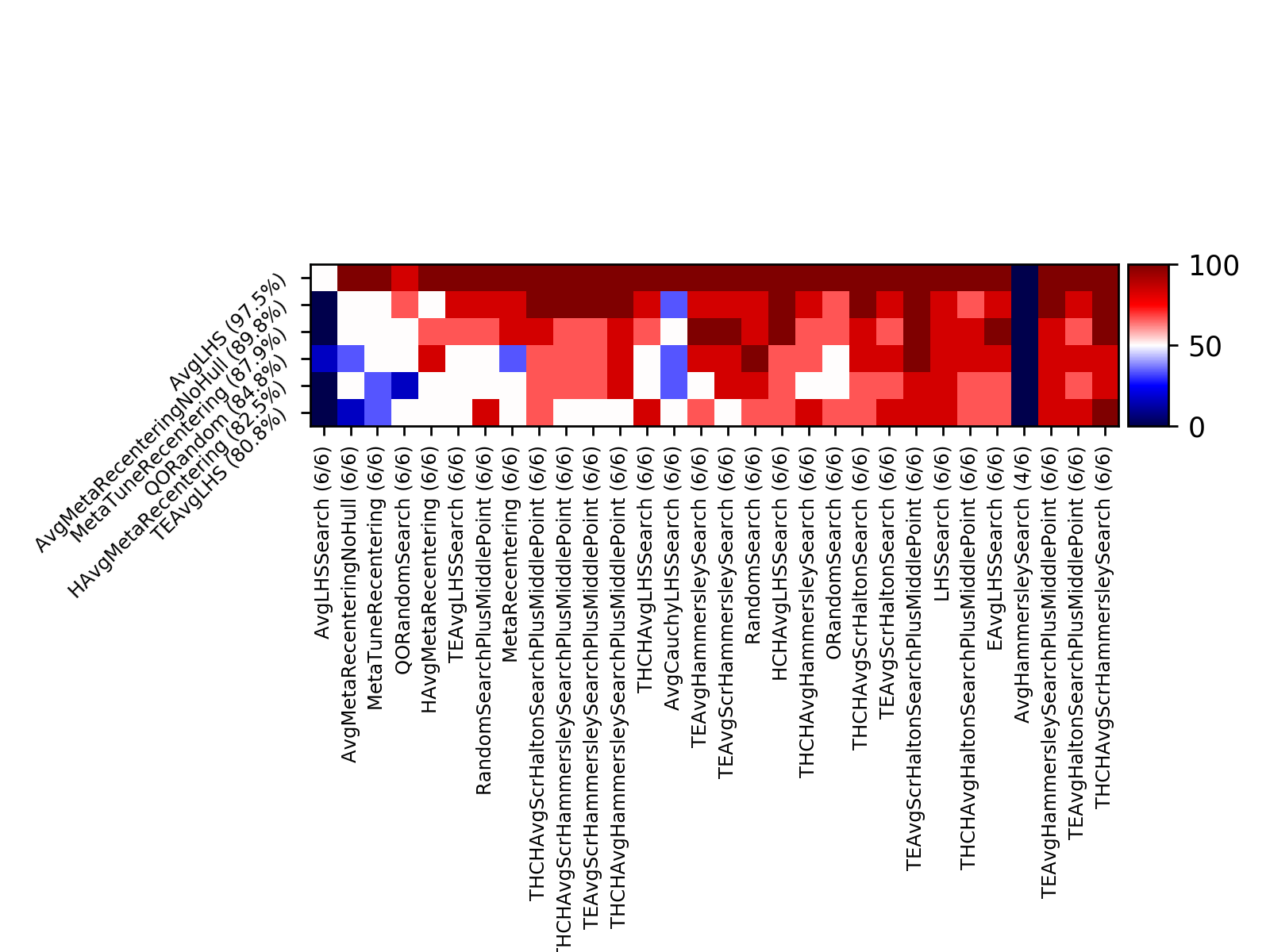}
\includegraphics[trim={10 10 12 80}, clip,width=.32\textwidth]{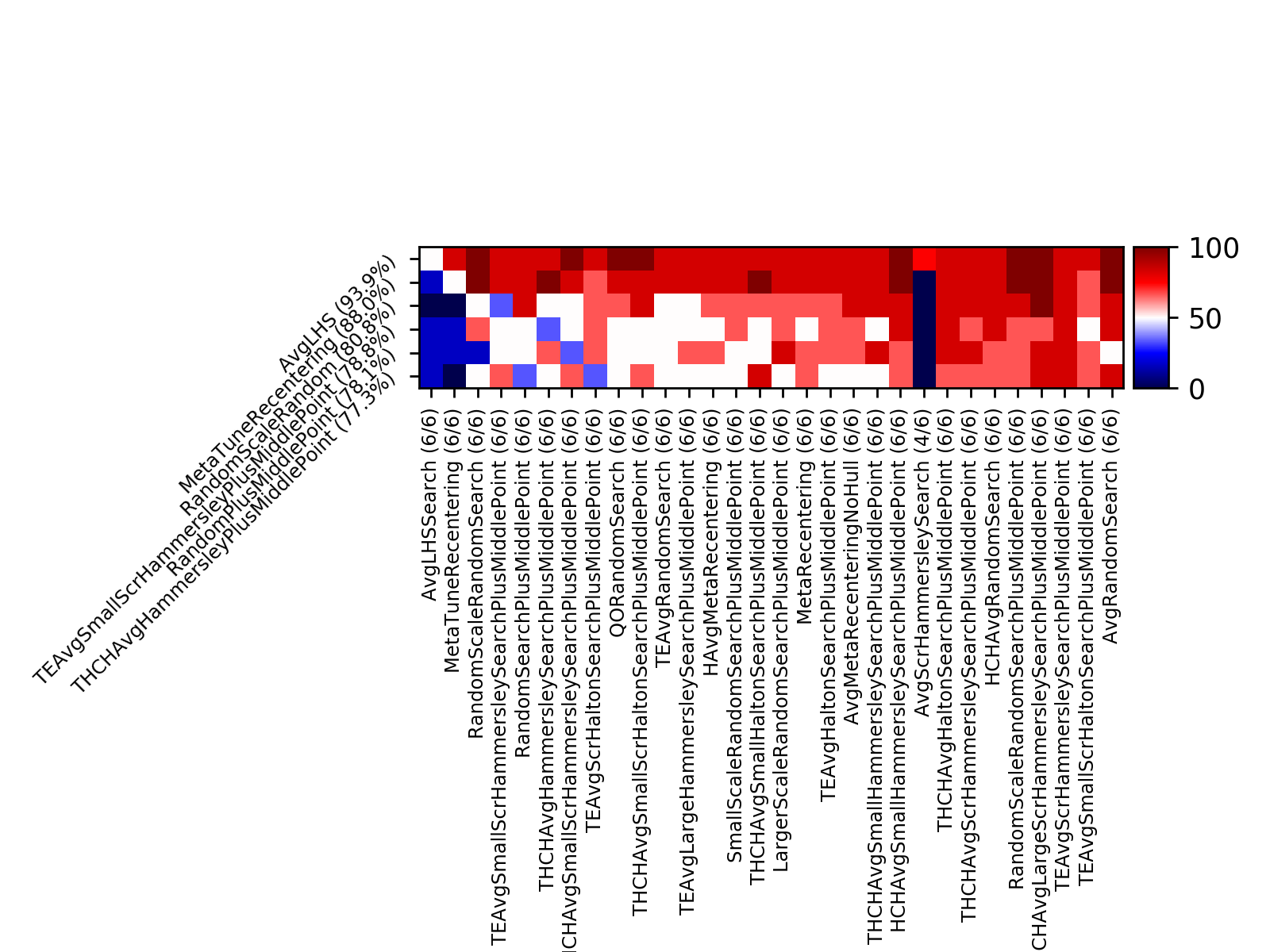}
\includegraphics[trim={10 10 12 80}, clip,width=.32\textwidth]{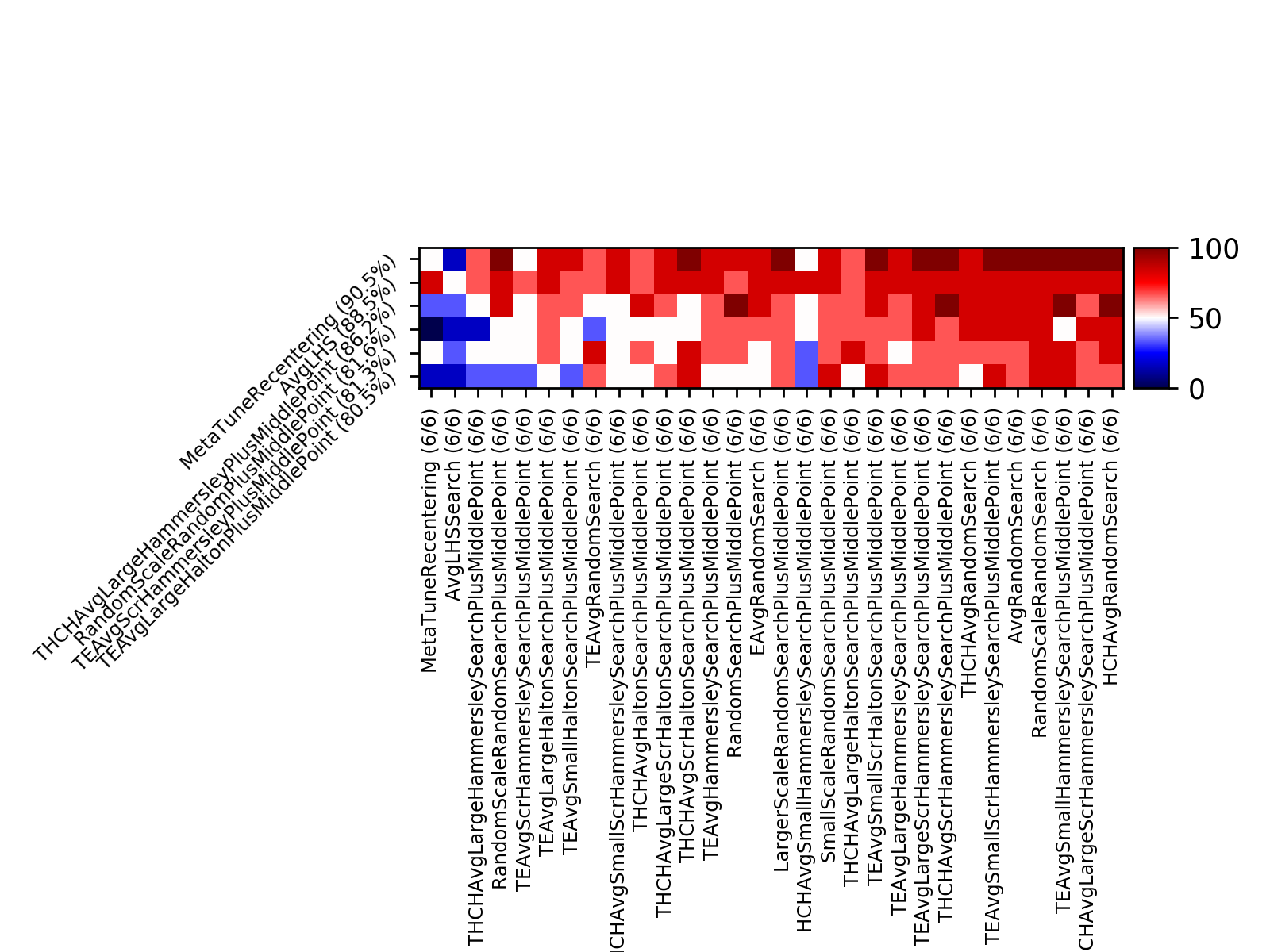}   \\
~\hfill Dimension $20$ \hfill ~ \hfill Dimension $200$ \hfill ~ \hfill Dimension $2000$\hfill ~\\
\caption{Results on the sphere function, per dimensionality. Results are still averaged over 6 values of the budget, namely $30$, $100$, $3000$, $10000$, $30000$, $100000$. \otc{Our method becomes better and better as the dimension increases.}}
    \label{toto2}
\end{figure}
\begin{figure}[t]
    \centering
\includegraphics[width=.44\textwidth]{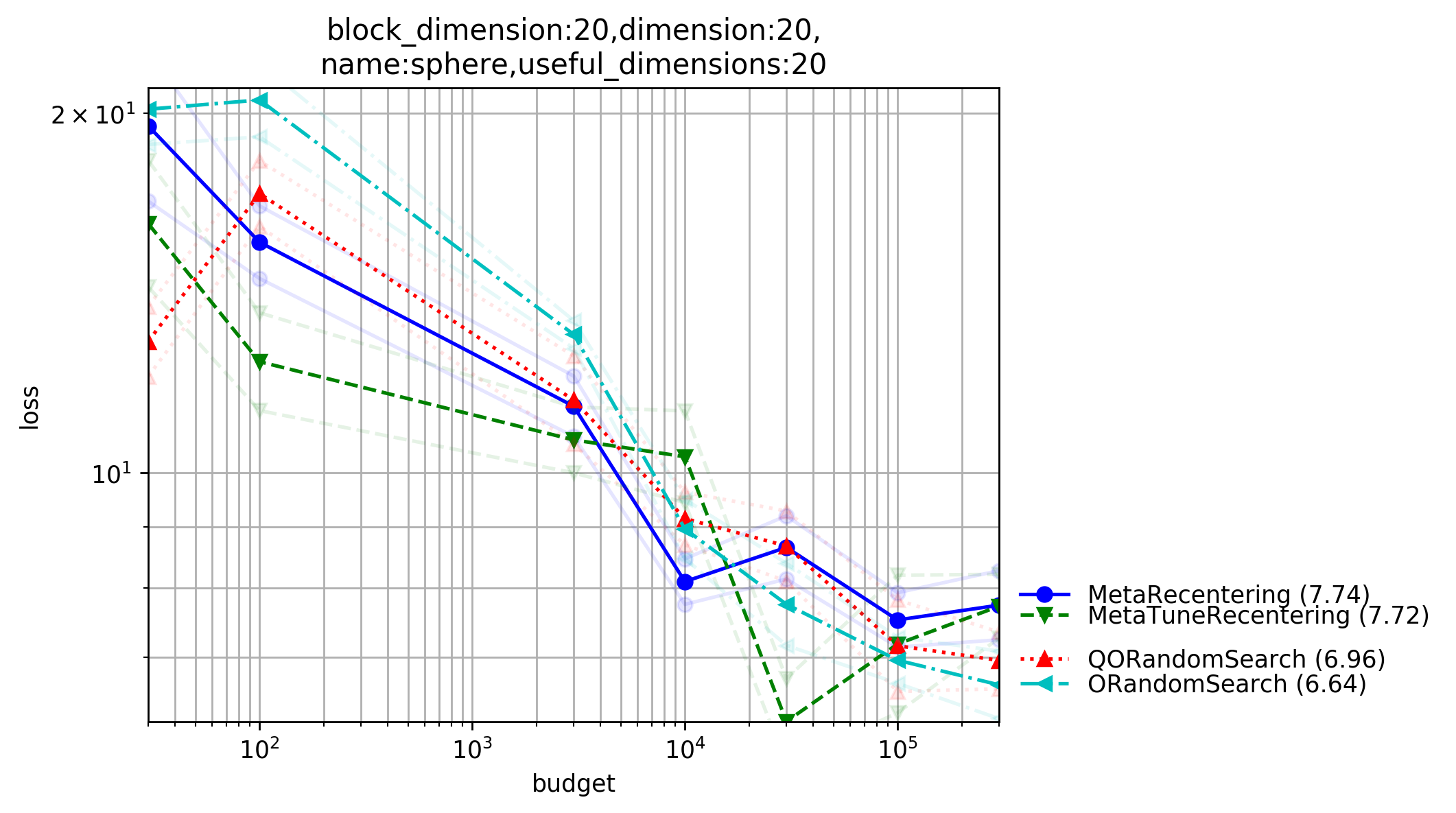}
 \includegraphics[width=.44\textwidth]{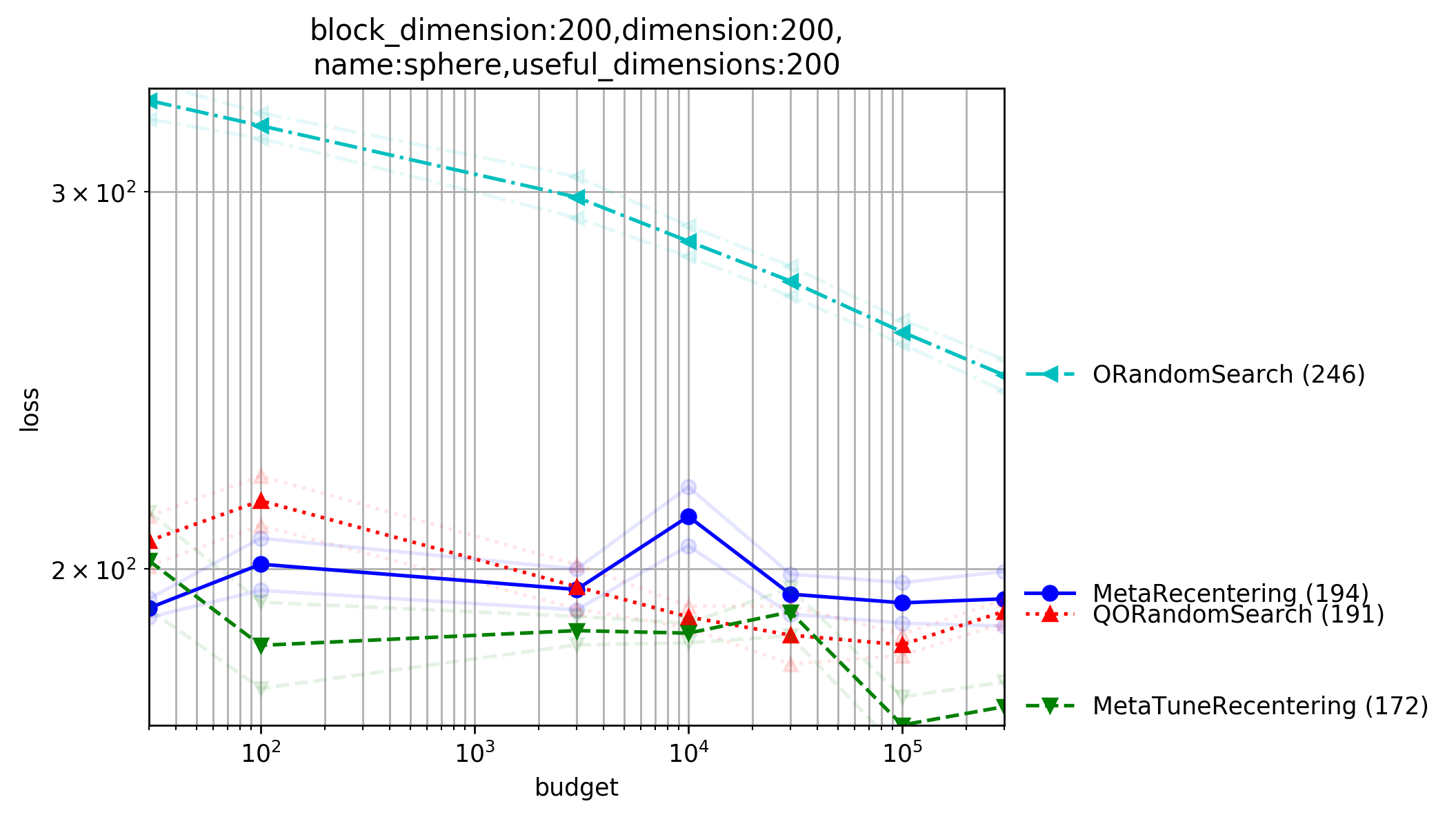}\\
 ~\hfill Dimension $20$ \hfill ~ \hfill Dimension $200$ \hfill ~ \\
  \includegraphics[width=.44\textwidth]{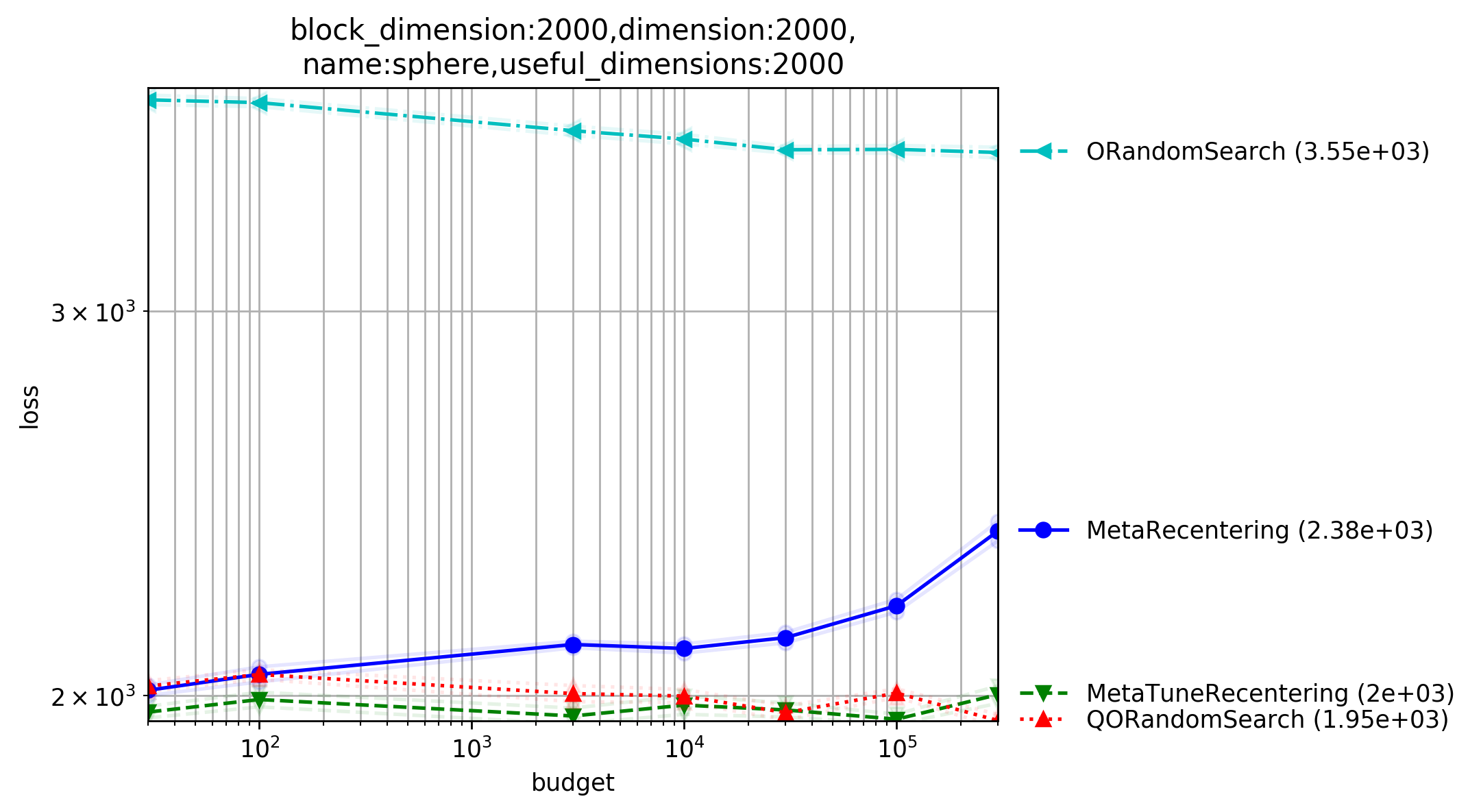}     \includegraphics[width=.44\textwidth]{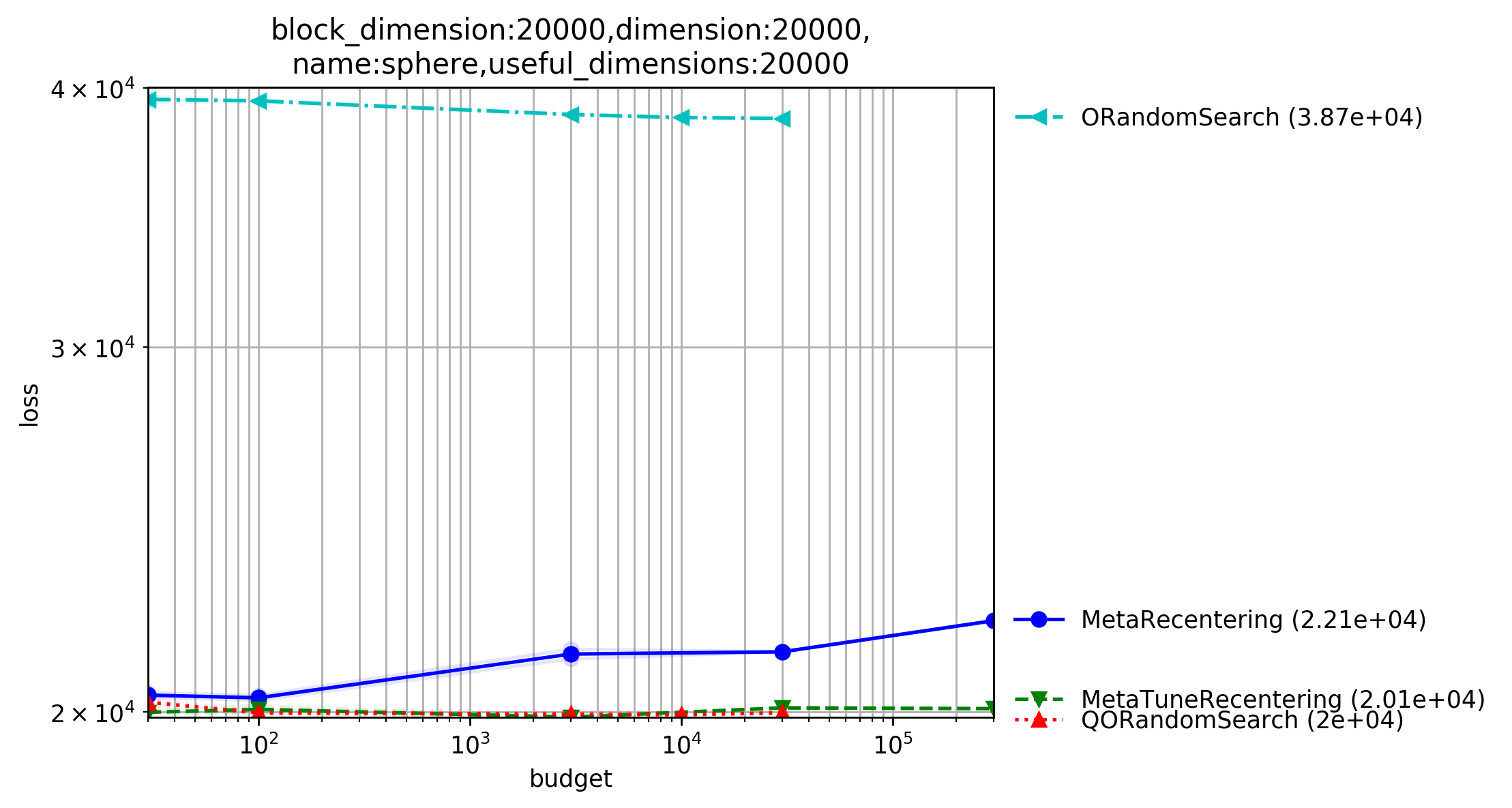}\\
 ~\hfill Dimension $2000$ \hfill ~ \hfill Dimension $20000$ \hfill ~ \\  
    \caption{Same context as Fig. \ref{toto2}, with $x$-axis = budget and $y$-axis = average simple regret. We see the failure of \texttt{MetaRecentering} in the worsening performance as budget goes to infinity: the budget has an impact on $\sigma$ which becomes worse, hence worse overall performance. We note that quasi-opposite sampling can perform decently in a wide range of values. Opposite Sampling is not much better than random search in high-dimension. \carola{Our \texttt{MetaTuneRecentering} shows decent performance: in particular, simple regret decreases as $\lambda\to\infty$.}}
    \label{toto2bis}
\end{figure}
\carola{Motivated by the significant improvements presented above, we now investigate whether the advantage of our rescaling factor translates to other optimization tasks. To this end, we first analyze a DoE setting, in which an underlying (and typically not explicitly given) function $f$ is to be minimized through a parallel evaluation of $\lambda$ solution candidates $x_1, \ldots, x_{\lambda}$, and regret is measured in terms of $\min_i f(x_i) - \inf_x f(x)$. In the broader \olivier{machine learning} literature, and in particular in the context of hyper-parameter optimization, this setting is often referred to as \textit{one-shot optimization}~\cite{bergstra,icmldoe}.}
\paragraph{Experimental Setup.} All our experiments are implemented and freely available in the Nevergrad platform~\cite{nevergrad}.  
Results are presented as shown in Fig.~\ref{doexp}. Typically, the six best methods are displayed as rows. \olivier{The 30 best performing} methods are presented as columns. The order for rows and for columns is the same: algorithms are ranked by their average winning frequency, measured against all other algorithms in the portfolio. 
% The heatmaps show the winning frequency for a given method $x$ \carola{(row)} against method $y$ \carola{(column)}, averaged over all settings and all replicas \olivier{(i.e. random repetitions)}, i.e., the fraction of runs in which algorithm $x$ outperformed algorithm $y$.
The heatmaps show \olivier{the fraction} of runs in which algorithm $x$ \carola{(row)} outperformed algorithm $y$ \carola{(column)}, averaged over all settings and all replicas (i.e. random repetitions).
The settings are typically sweepings over various budgets, dimensions, and objective functions.\footnote{\carola{Detailed results for individual settings are available at \url{http://dl.fbaipublicfiles.com/nevergrad/allxps/list.html}.}}%\carolanotetwo{Olivier, pls check if you want to have the reference to that webpage...} 
\carola{The numbers in the captions of the columns indicate the number of  settings for which the algorithms are compared against each other. That is, a bracket ``(6/6)'' is to be read as ``the winning frequencies are averaged over all six out of a total number of six settings''. For each tested (algorithm, problem) pair 20 independent runs are performed: a (6/6) case is thus based on a total number of 120 runs.}
\paragraph{Algorithm Portfolio.}
Several rescaling methods are already available on Nevergrad. A large fraction of these have been implemented by the authors of~\cite{icmldoe}; in particular:
\begin{itemize}
    \item The replacement of one sample by the center. These methods are named ``\texttt{midpointX}'' or ``\texttt{XPlusMiddlePoint}'', where \texttt{X} is \olivier{the original method that has been modified that way.} 
    \item The rescaling factor \texttt{MetaRecentering} empirically derived in~\cite{icmldoe}: $\sigma = \frac{1+\log(\lambda)}{4\log(d)}$.
    %presented in Eq.~\eqref{metarctg} below. 
    %This factor has been empirically derived in~\cite{icmldoe}.%, a step-size $\sigma$ as in.
    %\carolanote{which name does this one have? Should be mentioned here}
    \item The quasi-opposite methods suggested in~\cite{centerbased}, with prefix ``\texttt{QO}''\olivier{: when $x$ is sampled, then another sample $c-rx$ is added, with $r$ uniformly drawn in $[0,1]$ and $c$ the center of the distribution.}
    \end{itemize}\carola{\olivier{We also include} in our comparison \olivier{a different type of one-shot optimization techniques, independent of the present work, currently available in the platform:} they use the information obtained from the sampled points to recommend a point $x$ that is not necessarily one of the $\lambda$ evaluated ones. These \textit{``one-shot+1''} strategies \olivier{have the prefix} ``\texttt{Avg}''. }%\olivier{Remarkably,} our rescaled sampling still outperforms these more powerful methods.}
We keep all these and all other sampling strategies available in Nevergrad for our experiments.   %\otc{The methods containing ``Avg'' in the name are different by nature: they find an approximation of the optimum using the sampled points (assuming a model of the fitness values), but do not have to choose a point among the $X_1,\dots,X_\lambda$: we note that we still outperform those methods.}
We add to this existing Nevergrad portfolio our own rescaling strategy, which uses the scaling factor derived in Sec.~\ref{sec:theory}; i.e.,  $\sigma = \sqrt{\log(\lambda)/d}$. We refer to this sampling strategy as \texttt{MetaTuneRecentering}, defined below.
Both scaling factors \texttt{MetaRecentering}~\cite{icmldoe} and \texttt{MetaTuneRecentering} (our equations) are applied to quasirandom sampling (more precisely, scrambled Hammersley~\cite{hammersley,atanasov}) rather than random sampling. We provide detailed specifications of these methods and the most important ones below, whereas \olivier{we skip the dozens of other methods: they are open sourced in Nevergrad~\cite{nevergrad}.}
\paragraph{From $[0,1]^d$ to Gaussian quasi-random, random or LHS sampling:}
\olivier{Random sampling, quasi-random sampling, Latin Hypercube Sampling (or others) have a well known definition in $[0,1]^d$ (for quasi-random, see Halton~\cite{halton60} or Hammersley~\cite{hammersley}, possibly boosted by scrambling~\cite{atanasov}; for LHS, see \cite{mckay}). To extend to multidimensional Gaussian sampling, we use that if $U$ is a uniform  random variable on $[0,1]$ and $\Phi$ the standard Gaussian CDF, then $\Phi^{-1}(U)$ simulates a $\mathcal{N}(0,1)$ distribution. We do so on each dimension: this provides a Gaussian quasi-random, random or LHS sampling. }

\olivier{Then, one can rescale the Gaussian quasi-random sampling with the corresponding factor $\sigma$ for \texttt{MetaRecentering} ($\sigma = \frac{1+\log(\lambda)}{4\log(d)}$~\cite{icmldoe}) and \texttt{MetaTuneRecentering} ($\sigma=\sqrt{\log(\lambda)/d}$): for $i\leq \lambda$ and $j\leq d$, $x_{i,j}=\sigma\phi^{-1}(h_{i,j})$ where $h_{i,j}$ is the $j^{th}$ coordinate of a $i^{th}$ Scrambled-Hammersley point.} %Then, \texttt{MetaTuneRecentering} corresponds to a scrambled-Hammersley sampling, with rescaling $\sigma=\sqrt{\log(\lambda)/d}$.
%}
% \olivier{The general principle is as follows:
% \begin{itemize}
%     \item Random sampling or quasi-random sampling or Latin Hypercube Sampling (or others) have a well known definition in $[0,1]^d$. Nevergrad proposes several quasi-random samplings, including Halton\cite{halton1960}, Hammersley\cite{hammersley}, possibly boosted by scrambling\cite{atanasov}.
%     \item This sampling is translated into a Gaussian random sampling (or quasi-random sampling, etc) using cumulative distribution functions. This corresponds to ${\cal N}(0,I_d)$ in the random Gaussian case: using other samplings leads to a natural definition of quasi-random Gaussian sampling or Latin Hypercube Gaussian Sampling.
%     \item Possibly, we multiply by a rescaling factor $\sigma$.
%     \item If the target domain is $[0,1]^d$ we can then come back using, again, cumulative distribution functions.
% \end{itemize}
% MetaTuneRecentering corresponds to a scrambled-Hammersley sampling, with rescaling $\sigma=\sqrt{\log(\lambda)/d}$.
% }
%\carolanote{Olivier, is this a good reference, i.e., do you describe there what it means to superpose the scaling on the quasi-random sampling?}
% \subsubsection{Design of experiments (DOE): the testbed in Nevergrad.}
% We compare the simple regret of various one-shot optimization methods. Results are presented in Fig.~\ref{doexp}.
\paragraph{Results for the Full DoE Testbed in Nevergrad.} 
\carola{Fig.~\ref{doexp} displays aggregated results for the Sphere, the Cigar, and the Rastrigin functions, for three different dimensions and six different budgets. We observe that our \texttt{MetaTuneRecentering} strategy performs best, with a winning frequency of 80\%. It positively compares against all other strategies from the portfolio, with the notable exception of \texttt{AvgLHS}, which, in fact, compares favorably against every single other strategy, but with a lower average winning frequency of 73.6\%. Note here that \texttt{AvgLHS} is one of the \textbf{``oneshot+1''} strategies, i.e., it has not only one more sample, but it is also allowed to sample its recommendation adaptively, in contrast to our fully parallel \texttt{MetaTuneRecentering} strategy. It performs poorly in some cases (Rastrigin) and does not make sense as an initialization \olivier{(Sect. \ref{sec:initialization})}. } 
\paragraph{Selected DoE Tasks.} 
Figs.~\ref{toto3} breaks down the aggregated results from Fig.~\ref{doexp} by the three different functions. From this figure we see that \texttt{MetaTuneRecentering} scores second on sphere (where \texttt{AvgLHS} is winning) , third on Cigar (after \texttt{AvgLHS} and \texttt{QORandom}), and first on Rastrigin. This fine performance is quite remarkable, given that the portfolio contains quite sophisticated and highly tuned methods. \olivier{In addition, the {\texttt{AvgLHS}} methods, sometimes performing better on the sphere, besides using more capabilities than we do as it is a ``oneshot+1'' method, had poor results \olivier{for Rastrigin} (not even in the 30 best methods).} On sphere, the difference to the third and following strategies is significant (87.3\% winning rate against 77.5\% for the next runner-up). On Cigar, the differences between the first four strategies are greater than 4 percentage points each, whereas on Rastrigin the average winning frequencies of the first five strategies is comparable, but significantly larger than that of the sixth one (which scores 78.8\% against $>$94.2\% for the first five DoEs). 
Fig.~\ref{toto} zooms into the results for the sphere function, and breaks them further down by available budget $\lambda$ (note that the results are still averaged over the three dimensions $20$, $200$, $2000$). \texttt{MetaTuneRecentering} scores second in all six cases. 
A breakdown of the results for sphere by dimension (and aggregated over the six available budgets) is provided in Fig.~\ref{toto2} and Fig.~\ref{toto2bis}. \olivier{For dimension $20$, we see that \texttt{MetaTuneRecentering} ranks third, but, interestingly, the two first methods are ``oneshot+1'' style (\texttt{Avg} prefix).}
%\texttt{AvgMetaRecenteringNoHull} variant (which is the \textit{one-shot+1} coounterpart of {\texttt{MetaRecentering}} using as final recommendation the average of the best \olivier{$\min(d,\lambda/4)$}} evaluated points) ranks second, whereas the already mentioned \texttt{AvgLHS} ranks first. 
%The difference between \texttt{MetaTuneRecentering} and \texttt{AvgMetaRecenteringNoHull} is quite small, with winning frequencies of 87.9\% against 89.8\%. 
In dimension $200$, \texttt{MetaTuneRecentering} ranks second, with considerable advantage over the third-ranked strategy (88.0\% vs. 80.8\%). Finally, for the largest tested dimension, $d=2000$, our method ranks first, with an average winning frequency of 90.5\%.  
\subsection{Rescaled Sampling for Better Initialization of Iterative Optimization Heuristics}\label{sec:initialization}
We now move from the one-shot settings considered thus far to \textit{iterative optimization}, and show that our scaling factor can also be beneficial in this context. More precisely, we analyze the impact of initializing efficient global optimization (EGO~\cite{ego}, a special case of Bayesian optimization) and differential evolution (DE~\cite{de}) by a population that is sampled from a distribution that uses our variance scaling scheme. 
It is well known that a proper initialization can be very critical for the performance of these solvers;  see~\cite{feurer2015initializing,inoculation2,centerbased,qrinit,BossekDK20} for discussions. %, and several initialization schemes have been proposed in the literature. 
\begin{figure}[tp]
    \centering
    \includegraphics[trim={10 10 12 80}, clip,width=.44\textwidth]{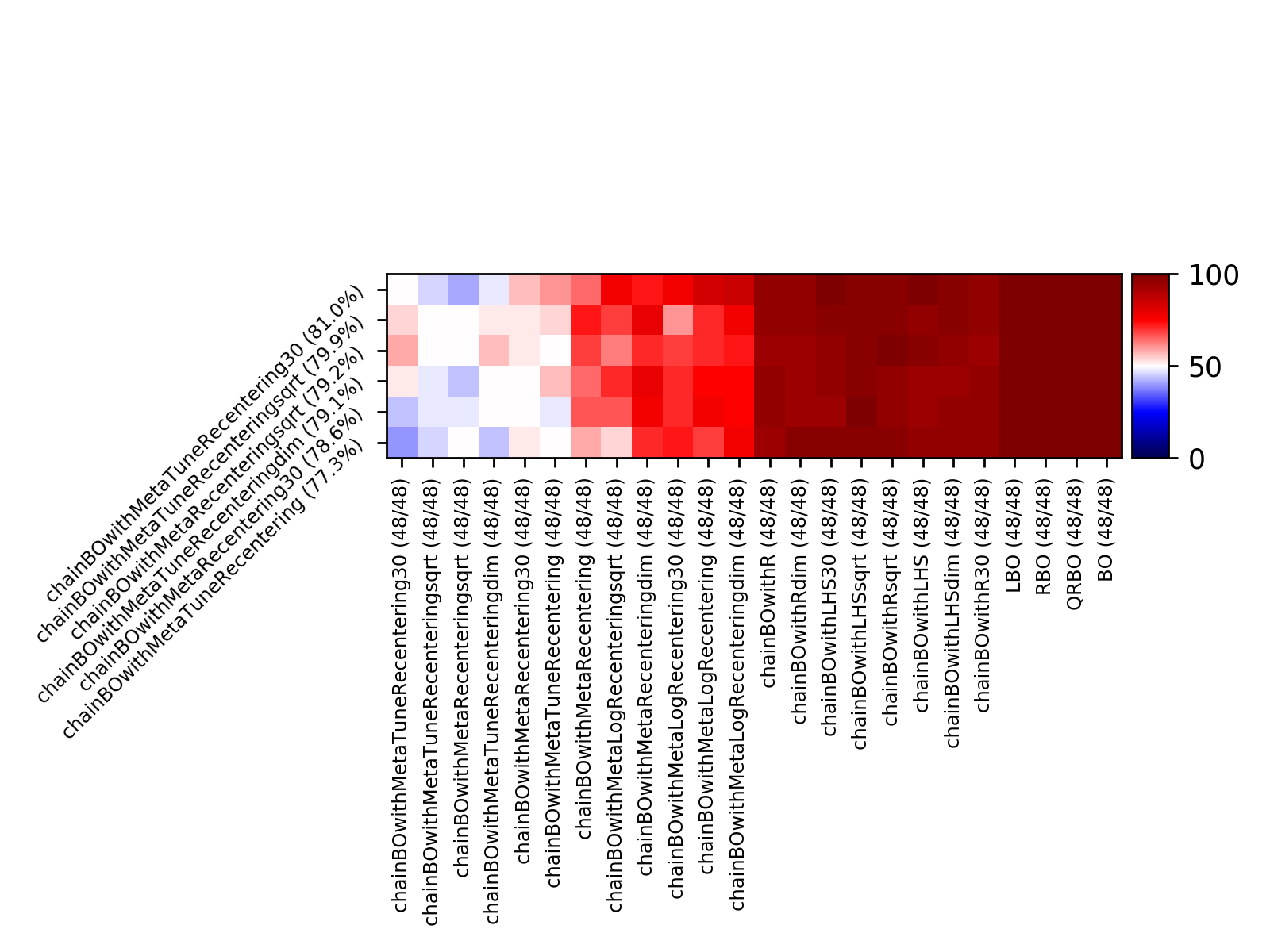}
    \includegraphics[trim={10 10 12 80}, clip,width=.44\textwidth]{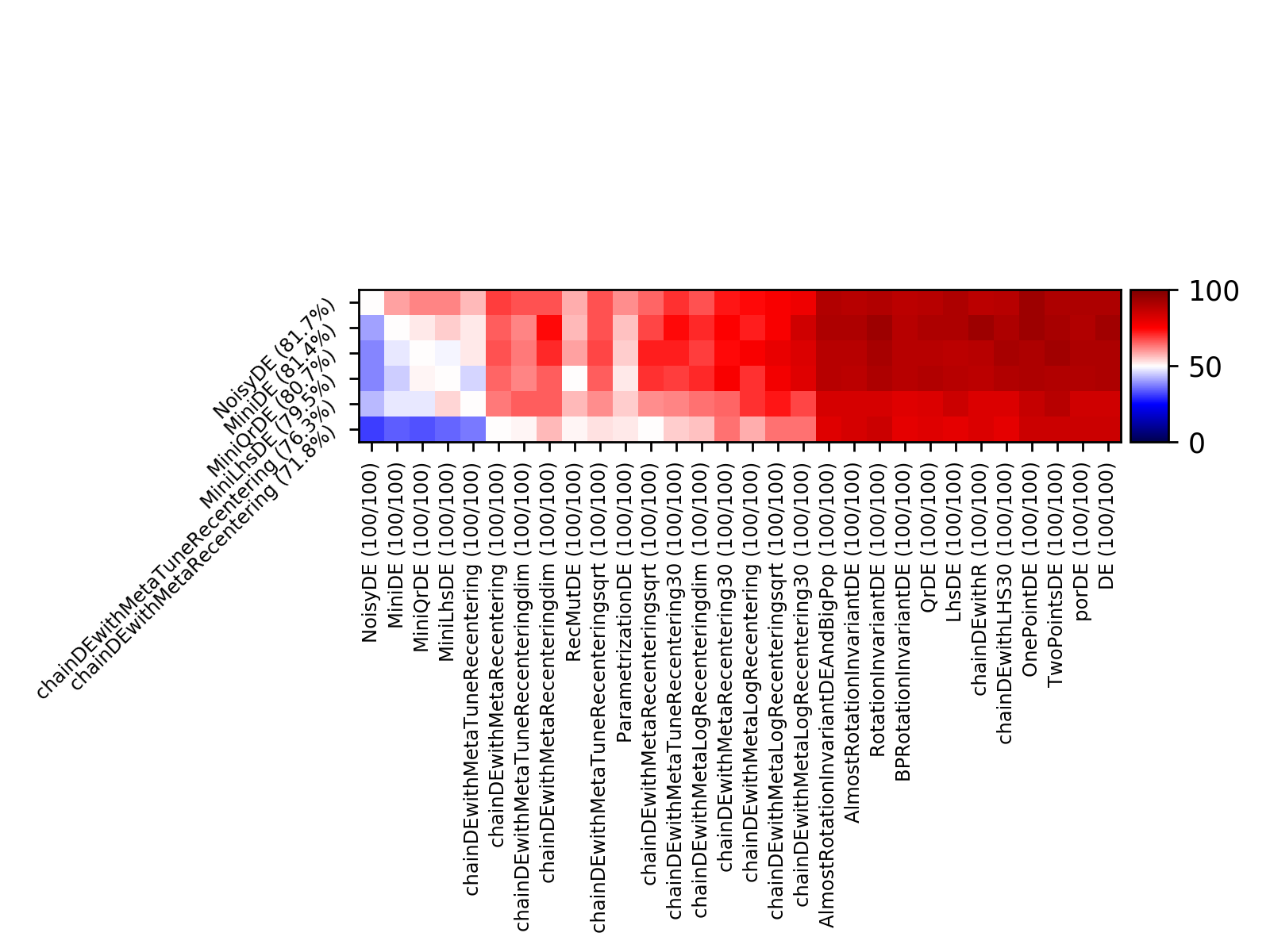}
    \caption{Performance comparison of different strategies to initialize Bayesian Optimization (BO, left) and Differential Evolution (DE, right). A detailed description is given in Sec.~\ref{sec:initialization}. \otc{\texttt{MetaTuneRecentering} performs best as an initialization method. In the case of DE, methods different from the traditional DE remain the best on this testcase: when we compare DE with a given initialization and DE initialized with \texttt{MetaTuneRecentering}, \texttt{MetaTuneRecentering} performs best in almost all cases. }}
    \label{deinitxp}\label{boinit}
\end{figure}
Fig.~\ref{boinit} summarizes the results of our experiments. As in the previous setups, we compare against existing methods from the Nevergrad platform, to which we have just added our rescaling factor termed \texttt{MetaTuneRecentering}. For each initialization scheme, \carola{four} different initial population sizes are considered: denoting by $d$ the dimension, by $w$ the  parallelism \carola{(i.e., the number of workers)}, and by $b$ the total budget that the algorithms can spend on optimizing the given optimization task, the initial population $\lambda$ is set as   
$\lambda=\sqrt{b}$ for \texttt{Sqrt}, as $\lambda=d$ for \texttt{Dim}, $\lambda=w$ for no suffix, and as $\lambda=30$ when the suffix is \texttt{30}. 
% We recall that our variance scaling from  Eq.~\eqref{metatunerecentering} is termed \texttt{MetaTuneRecentering}, and that 
As in Sec.~\ref{rg} we \carola{superpose} our scaling scheme on top of the quasi-random Scrambled Hammersley sequence suggested in~\cite{icmldoe}, but we also consider random initialization rather than quasi-random (indicated by the suffix ``\texttt{R}'') and Latin Hypercube Sampling~\cite{mckay} (suffix ``\texttt{LHS}''). 
% versions for choosing the  number $\lambda$ of points in the initialization depending on the dimension $d$, the degree of parallelism $w$, and the budget $b$: $\lambda=\sqrt{b}$  for Sqrt, $\lambda=d$ for Dim, $\lambda=w$ for no suffix, $\lambda=30$ when the suffix is 30. 
% \texttt{MetaTuneRecentering} is our scaling as in Eq.~\eqref{metatunerecentering}, combined with  Scrambled Hammersley, i.e. the recommended quasirandom method used in~\cite{icmldoe}. \texttt{MetaRecentering} is the previous state of the art in the platform, using Eq.~\ref{metarctg} as a scaling.
The left chart in Fig.~\ref{boinit} is for the Bayesian optimization case. It aggregates results for 48 settings, which stem from Nevergrad's ``parahdbo4d'' suite. It comprises the four benchmark problems Sphere, Cigar, Ellipsoid and Hm. Results are averaged over the total budgets $b\in \{25, 31, 37, 43, 50, 60\}$, dimension $d\in\{20,2000\}$, and parallelism $w=\max(d,\lfloor b/6\rfloor)$. The parallelism is $20$. 
We observe that a BO version using our \texttt{MetaTuneRecentering} performs best, and that several other variants using this scaling appear among the top-performing configurations. 
The chart on the right of Fig.~\ref{boinit} summarizes results for Differential Evolution. Since DE can handle larger budgets, we consider here a total number of 100 settings, which correspond to the testcase named ``paraalldes'' in Nevergrad. In this suite, results are averaged over budgets $b \in \{10, 100, 1000, 10000, 100000\}$, dimensions $d\in \{5, 20, 100, 500, 2500\}$, parallelism $w=\max(d,\lfloor b/6\rfloor)$, and again the objective functions Sphere, Cigar, Ellipsoid, and Hm. The parallelism is $20$. Specialized versions of DE perform best for this testcase, but we see that DE initialized with our \texttt{MetaTuneRecentering} strategy ranks fifth \olivier{(outperformed only by ad hoc variants of DE)}, with an overall winning frequency that is not much smaller than that of the top-ranked \texttt{NoisyDE} strategy (76.3\% for \texttt{ChainDEwithMetaTuneRecentering} vs. 81.7\% for \texttt{NoisyDE}) - and almost always outperforms the rescaling used in the original Nevergrad. 
% We observe that our \texttt{MetaTuneRecentering} 
%scaling performs best in most cases as a one-shot optimization method (Fig. \ref{doexp}, \ref{toto}, \ref{toto2}, and~\ref{toto3}). This scaling also 
% performs quite well as an initialization method for optimization algorithms. \carolanote{We should add a few more details about the results}
%(Fig.~\ref{boinit}).
% \otc{With just this rescaling, we perform quite well also compared to methods building an approximate optimum $\hat X$ using a model of the objective function (methods with ``Avg'' in the name) - whereas we do not have to spend time validating our approximate optimum given that it is one of the evaluated points $X_1,\dots,X_\lambda$.}
%\subsection{The multi-epochs case}
%TODO discuss choice of values in the state of the art
%
%TODO experimental results
\section{Conclusions and Future Work}
\carola{
We have investigated \olivier{the scaling} of the variance of random sampling in order to minimize the expected \olivier{regret.} While previous work~\cite{icmldoe} had already shown that the optimal scaling factor is not identical to that of the prior distribution from which the optimum is sampled (unless the sample size is exponentially large in the dimension), it did not answer the question how to scale the variance optimally. In this work, we have proven that standard \olivier{deviations} scaled as $\sigma=\sqrt{\log(\lambda)/d}$ gives, with probability at least $1/2$, a sample that is significantly closer to the optimum than \olivier{the previous known strategies}. We have also shown that the gain achieved by our rescaled sampling strategy is asymptotically optimal. Moreover, we have shown that any decent scaling factor is asymptotically at most as large as our proposed one.}
\carola{The empirical assessment of our rescaled sampling strategy confirms decent performance not only on the sphere function, but also on other classical benchmark problems. We have furthermore given indication that the sampling might help improve state-of-the-art numerical heuristics based on differential evolution or using Bayesian surrogate models. \olivier{Our proposed one-shot method performs best in many cases, sometimes outperformed by e.g. \texttt{AvgLHS}, but is stable on a wide range of problems and meaningful also as an initialization method (as opposed to \texttt{AvgLHS})}.}
\carola{Whereas our theoretical results can be extended to quadratic forms \olivier{(by conservation of barycenters through linear transformations)}, an extension to wider families of functions (e.g., families of functions with order 2 Taylor expansion) is not straightforward. Apart from extending our results to broader function classes, another direction for future work comprises extensions to the multi-epoch case. Our empirical results on DE and BO gives first indication that a properly scaled variance can also be beneficial in iterative sampling. \olivier{Note, however, that in the latter case, we only adjusted the initialization}, not the later sampling steps. This forms another promising direction for future work.}
% We prove a rescaling factor (Eq.~\ref{metatunerecentering}) for the sampling of a population. This validates empirical rules of thumb such as~\cite{hansen2006cma,centerbased,icmldoe}, with a more precise rule. This rule is validated in experimental results using an open source platform\cite{nevergrad}.
% {\bf{Further work.}}
% Our results can be extended easily to quadratic forms (by conservations of barycenters through linear transformations) but the extension to wider families of functions (e.g. families of functions with order 2 Taylor expansion) is not straightforward. In addition, we did not prove anything for the multi-epoch case.

\vspace{1ex}
\textbf{Acknowledgements.} This work was initiated at Dagstuhl seminar 19431 on Theory of Randomized Optimization Heuristics. 
%\oliviernote{Bibliography should be fixed, there are dirty things in it.}
\newpage

\bibliographystyle{splncs04}
\bibliography{lsca,biblio}%,doe,bibliodoe,bibliolsca,biblio,biblio2}

%%%%%%%%%%%%%%%%%%%%%%%%%%%%%%%%%%%%%%%%%%
%%%%%%%%%%%%    APPENDIX   %%%%%%%%%%%%%%
%%%%%%%%%%%%%%%%%%%%%%%%%%%%%%%%%%%%%%%%%%
\newpage

\section*{Appendix A: Relevant Concentration Bounds for $\chi^2$ Distributions}
\label{back}

We recall some basic definitions and properties of the central and the non-central $\chi^2$ distributions, which are needed in the proofs of Theorems~\ref{thm:suff} and~\ref{thm:nec}. %\carolanote{For reasons of space, we can only sketch these here; the interested reader is referred to~\cite{SomeBook} for a more extended discussion.@Laurent, do you happen to know a good textbook?} 

\begin{definition}{(Central $\chi^2$-distribution)}
Let $X_1,...,X_d$ be $d$ independent random variables drawn from the standard normal distribution $\mathcal{N}(0,1)$. Then the random variable $U = X_1^2+...+X_d^2$ follows a central $\chi^2(d)$ distribution with $d$ degrees of freedom. 
\end{definition}
As mentioned previously, the squared distance $||x^*||^2$ of $x^*$ to the middle point $0$ follows the central $\chi^2(d)$ distribution. This is thus also the distribution of the performance of the random sampling strategy using $\sigma^2=0$. In our proofs we will make use of the following properties of this distribution. 
%e central $\chi^2$ distribution. 
% \carola{If we run short of space, we can omit the properties environment} 
\begin{property}{(Properties of $\chi^2$ distribution)}
\label{ppty:chi2}
% Let $U\sim \chi^2(d)$.
% \begin{enumerate}
%     \item $\mathbb{E}(U)=d$, $\text{var}(U) = 2d$.
%     %\item The density of a $\chi^2(d)$ distribution is $p_U(x)=\frac{1}{2^{k/2}\Gamma(k/2)}x^{k/2-1}e^{-x/2}$, for all $x\geq0$.\carolanote{Do we need this?}
%     \item \textbf{Concentration bound:} For all $t\in[0,1]$ it holds that  $\mathbb{P}\left[\lvert\frac{U}{d}-1\rvert\geq t\right]\leq 2e^{-\frac{dt^2}{8}}$.
% \end{enumerate}
Let $U\sim \chi^2(d)$. Then $\mathbb{E}(U)=d$, $\text{var}(U) = 2d$, and for all $t\in[0,1]$ it holds that $\mathbb{P}\left[\lvert\frac{U}{d}-1\rvert\geq t\right]\leq 2\exp({-\frac{dt^2}{8}})$.
\end{property}

While the central $\chi^2$ distribution suffices for the analysis of the middle point sampling strategy, \textit{non-central $\chi^2$ distribution} are required in the analysis of our Gaussian sampling with rescaled variance.
%The definition of the $\chi^2$ distributions can be extended to the non-zero mean normal case. This is the \textit{non-central $\chi^2$ distribution}, defined as follows.
\begin{definition}{(Non-central $\chi^2$-distribution)}
Let $X_1,...,X_d$ be independently drawn random variables satisfying $X_i \sim \mathcal{N}(\mu_i,1)$. 
Let $U = X_1^2+...+X_d^2$. 
The random variable $U$ follows a central $\chi^2(d,\mu)$ distribution with $d$ degrees of freedom and non-centrality parameter $\mu=\sum_{i=1}^d \mu_i^2$.
% \carolanote{$\mu$ was $\lambda$, but this is not good here, since we use $\lambda$ to denote the population size.}
\end{definition}
Note here that the non-central $\chi^2$ distribution only depends on $\sum_{i=1}^d \mu_i^2$, but not on the individual values $(\mu_1,...,\mu_d)$. 
Note further that, for a given point $x^* \in \R^d$, the distribution of the squared distance $||x - x^*||^2$ for $x \sim \N(0,I)$ follows the non-central $\chi^2(d,\mu)$ distribution with non-centrality parameter $\mu:=||x^*||^2$.

We recall some important properties of the non-central $\chi^2$ distribution.
% \carolanote{If we run short of space, we can omit the properties environment}  
\begin{property}{(Properties of the non-central $\chi^2$ distribution)}
\label{ppty:ncchi^2}
Let $U\sim \chi^2(d,\mu)$. Then $\mathbb{E}(U)=d+\mu$, $\text{var}(U) = 2(d+2\mu)$, and for any $\beta>1$ there exist positive constants $C_1$, $C_{\beta}$ such that for all $x\leq (\mu+d)/\beta$ it holds that 
\begin{equation}
\label{concentration1}
P\left(U\leq -x\right) \geq C_1 \exp{\left(-C_\beta \frac{x^2}{2\mu+d}\right)}.
\end{equation}
Moreover, for all $x>0$, it holds that 
\begin{equation}
    \label{concentration2}
P\left(U\leq -x\right) \leq \exp{\left(-\frac14 \frac{x^2}{2\mu+d}\right)}.
\end{equation}
% \begin{enumerate}
%     \item $\mathbb{E}(U)=d+\mu$, $\text{var}(U) = 2(d+2\mu)$.
% %    \item The density of a $U$ is $p_U(x)=\frac12 x^{k/2-1}e^{-(x+\mu)/2}\left(\frac{x}{\mu}\right)^{k/4-1/2}I_{k/2-1}(\sqrt{\mu x})$, for all $x\geq0$, where $I$ refers to a modified Bessel function of first kind. \carolanote{possible notation clash between $I_d$ from the normal distribution and this $I$ here. Can we use $B$ instead? Do we actually need this expression anywhere?}
%     \item \textbf{Concentration bounds from~\cite[Theorem~7]{chi2magical}.} For any $\beta>1$ there exist constants \carola{$C_1>0$ and $C_\beta>0$} 
%     %\carolanote{was:non-negative, but for $C_1=0$ the statement is trivial, so I replaced this by $>0$} 
%     such that for all $x\leq (\mu+d)/\beta$ it holds that 
% $$P\left(U\leq -x\right) \geq C_1 \exp{\left(-C_\beta \frac{x^2}{2\mu+d}\right)}.$$
% Moreover, for all $x>0$, it holds that  
% $$P\left(U\leq -x\right) \leq \exp{\left(-\frac14 \frac{x^2}{2\mu+d}\right)}.$$
% \end{enumerate}
\end{property}
Proofs for the concentration inequalities~\ref{concentration1} and~\ref{concentration2} can be found in~\cite[Theorem~7]{chi2magical}. 
\section*{Appendix B: Proof of Theorem~\ref{thm:suff} (Sufficient condition) }
\label{sec:suff}

We now present the proof of Theorem~\ref{thm:suff}, the sufficient condition for the scaling factor $\sigma^2$ to be beneficial over sampling the middle point. 
% \begin{proof}
% Let $\delta\in[1/2,1)$ and let $\lambda, d \in \mathbb{N}$. 
Let $\delta$, $\lambda$ and $d$ satisfy the conditions of Theorem~\ref{thm:suff}. 
Let $\epsilon,\sigma>0$. By the law of total probability it holds that, for all $t\leq 1$,  
\begin{align*}
   \mathbb{P}&\left[\min_{i\in[\lambda]}||x^*-x_i||^2 \leq \left(1-\epsilon\right)||x^*||^2\right] \\ &=\mathbb{P}\left[\min_{i\in[\lambda]}||x^*-x_i||^2 \leq \left(1-\epsilon\right)||x^*||^2 \mid \lvert\frac{||x^*||^2}{d}-1\rvert\leq t\right]
   \mathbb{P}\left[\lvert\frac{||x^*||^2}{d}-1\rvert\leq t\right]\\
   &+\mathbb{P}\left[\min_{i\in[\lambda]}||x^*-x_i||^2 \leq \left(1-\epsilon\right)||x^*||^2\big| \lvert\frac{||x^*||^2}{d}-1\rvert> t\right]
   \mathbb{P}\left[\lvert\frac{||x^*||^2}{d}-1\rvert> t\right].
\end{align*}
Eq.~\ref{eq:betterdelta} is therefore satisfied if 
\begin{eqnarray*}
\mathbb{P}\left[\min_{i\in[\lambda]}||x^*-x_i||^2 \leq \left(1-\epsilon\right)||x^*||^2\big| \lvert\frac{||x^*||^2}{d}-1\rvert\leq t\right]
   \mathbb{P}\left[\lvert\frac{||x^*||^2}{d}-1\rvert\leq t\right]\geq \delta.
% \label{eq:suff1}
\end{eqnarray*}
This equation, in turn, is satisfied if for all $y$ with $\lvert\frac{||y||^2}{d}-1\rvert\leq t$ it holds that
\begin{eqnarray}
\mathbb{P}\left[\min_{i\in[\lambda]}||x^*-x_i||^2 \leq \left(1-\epsilon\right)||x^*||^2\big|x^*=y\right]
\geq \frac{\delta}{\mathbb{P}\left[\lvert\frac{||x^*||^2}{d}-1\rvert\leq t\right]}.
\label{eq:suff2}
\end{eqnarray}
For the following computations, we fix $t:=d^{-1/3}$ and we set $\delta':=\delta/\mathbb{P}\left[\lvert\frac{||x^*||^2}{d}-1\rvert\leq t\right]$.

Let $x^*$ be such that $\lvert\frac{||x^*||^2}{d}-1\rvert\leq t$. Then, conditionally to $x^*$, we have 
\begin{align*}
  \mathbb{P}&\left[\min_{i\in[\lambda]}||x^*-x_i||^2 \leq \left(1-\epsilon\right)||x^*||^2\big| x^*\right] \\
  & = 1- \mathbb{P}\left[\min_{i\in[\lambda]}||x^*-x_i||^2 \geq \left(1-\epsilon\right)||x^*||^2\big|x^*\right]\\
  &=1-\mathbb{P}\left[||x-x^*||^2 \geq \left(1-\epsilon\right)||x^*||^2\big|x^*\right]^\lambda\\
  &=1-\left(1-\mathbb{P}\left[||x-x^*||^2 \leq \left(1-\epsilon\right)||x^*||^2\big|x^*\right]\right)^\lambda
\end{align*}
for an $x$ is distributed as a normal distribution $\N(0,\sigma^2 I)$. We recall that for such an $x$ the distribution of the term $||x-x^*||^2/\sigma^2$ (for fixed $x^*$) follows the non-central $\chi^2(d,\mu)$ distribution with non-centrality parameter $\mu:=||x^*||^2/\sigma^2$. We therefore obtain (through simple algebraic manipulations) that condition~\eqref{eq:suff2} holds if and only if 
\begin{align*}
% &\mathbb{P}\left[\min_{i\in[\lambda]}||x^*-x_i||^2 \leq \left(1-\epsilon\right)||x^*||^2\big| x^*\right]\geq \delta'\\
% \iff&
\mathbb{P}\left[U \leq \left(1-\epsilon\right)\frac{||x^*||^2}{\sigma^2}\right]\geq1-(1-\delta')^{1/\lambda}\,,
\end{align*}
with $U \sim \chi^2(d,\mu)$. % the non-central $\chi^2$ distribution with $d$ degrees of freedom with parameter $\frac{||x^*||^2}{\sigma^2}$. 
Let $Y  := U-\left(\frac{||x^*||^2}{\sigma^2}+d\right)$.
Then the previous condition is equivalent to
\begin{equation*}
    \mathbb{P}\left[Y \leq -\left(\epsilon\frac{||x^*||^2}{\sigma^2}+d\right)\right]\geq1-(1-\delta)^{1/\lambda}\,.
    % \label{ilovenumberedequations}
\end{equation*}

According to the concentration inequality~\ref{concentration1}, it holds that for any $\beta>1$, there exist constants $C_1>0$ and $C_\beta>0$ such that if 
\begin{equation}
\label{epscondi}
    \epsilon\frac{||x^*||^2}{\sigma^2}+d \leq \frac1\beta \left(\frac{||x^*||^2}{\sigma^2}+d\right),
\end{equation}
then 
% \begin{equation}
% (\epsilon\frac{||x^*||^2}{\sigma^2}+d)\leq \frac1\beta (\frac{||x^*||^2}{\sigma^2}+d)  
%  \label{eq:assbeta}
% \end{equation}
% Therefore Eq.~\ref{ilovenumberedequations} becomes: 
$$\mathbb{P}\left(Y\leq -\left(\epsilon\frac{||x^*||^2}{\sigma^2}+d\right)\right) \geq C_1 \exp{\left(-C_\beta \frac{(\epsilon\frac{||x^*||^2}{\sigma^2}+d)^2}{2\frac{||x^*||^2}{\sigma^2}+d}\right)}.$$
We deduce a sufficient condition for~\eqref{eq:suff2}, by noting that it is satisfied if, for all $x^*$ such that $\lvert\frac{||x^*||^2}{d}-1\rvert\leq t$, it holds that
\begin{equation}
\frac{\left(\epsilon\frac{||x^*||^2}{\sigma^2}+d\right)^2}{2\frac{||x^*||^2}{\sigma^2}+d} \leq A_\lambda,
\label{eq:suff3}
\end{equation}
with $A_\lambda := -\frac{1}{C_\beta}\left(\log\left(1-(1-\delta')^{1/\lambda}\right)-\log C_1\right)$. 

Let us now fix $\beta:=2$, $\epsilon := c_1\frac{\log\lambda}{d}$ and $\sigma^2:=c_2\frac{\log\lambda}{d}$, with $c_1 := \frac{1}{3 C_\beta}$ and $c_2:=c_1$. 
We show that, with these choices of $\beta$, $\epsilon$ and $\sigma$, inequalities~\eqref{epscondi} and~\ref{eq:suff3} are satisfied if $d$ is sufficiently large and $x^*$ satisfies $\lvert\frac{||x^*||^2}{d}-1\rvert\leq t$. 
% $\frac{\epsilon\frac{||x^*||^2}{\sigma^2}+d}{\frac{||x^*||^2}{\sigma^2}+d} \leq \frac1\beta$ and 
% $\frac{(\epsilon\frac{||x^*||^2}{\sigma^2}+d)^2}{2\frac{||x^*||^2}{\sigma^2}+d} \leq A_\lambda$ 
% are satisfied for such choice of $\beta$, $\epsilon$ and $\sigma$. We also recall that $t = d^{-1/3}$.
To this end, first note that 
$$\frac{\epsilon\frac{||x^*||^2}{\sigma^2}+d}{ (\frac{||x^*||^2}{\sigma^2}+d)}\leq  \frac{\frac{c_1}{c_2}(1+t)+1}{\frac{d}{c_2\log\lambda}(1-t)+1}.$$
Under the assumptions stated in~\eqref{ass3b} the term $\frac{\frac{c_1}{c_2}(1+t)+1}{\frac{d}{c_2\log\lambda}(1-t)+1}$ converges to zero as $d\rightarrow \infty$. 
We therefore obtain that, for $d$ sufficiently large and $x^*$ satisfying $\lvert\frac{||x^*||^2}{d}-1\rvert\leq t$, it holds that  $$\frac{\epsilon\frac{||x^*||^2}{\sigma^2}+d}{\frac{||x^*||^2}{\sigma^2}+d} \leq \frac1\beta,$$
which proves~\eqref{epscondi}. 

To show~\eqref{eq:suff3}, we first note that 
$$\frac{(\epsilon\frac{||x^*||^2}{\sigma^2}+d)^2}{2\frac{||x^*||^2}{\sigma^2}+d} \leq \frac{\left(\frac{c_1}{c_2}(1+t)+1\right)^2}{2\frac{d}{c_2\log\lambda}(1-t)+1}.$$ 
Under the assumptions stated in~\eqref{ass3b}, and since $d\rightarrow\infty$, we approximate $$\frac{\frac{c_1}{c_2}(1+t)+1}{\frac{d}{c_2\log\lambda}(1-t)+1}=\frac{c_2}{2}\left(\frac{c_1}{c_2}+1\right)^2\log \lambda+o(\log \lambda) = \frac{2}{3 C_\beta} \log \lambda+o(\log \lambda)$$ 
and $A_\lambda= \frac{1}{C_\beta}\log \lambda+o(\log\lambda)$, which shows that condition~\ref{eq:suff3} holds for $d$ sufficiently large and $x^*$ satisfying $\lvert\frac{||x^*||^2}{d}-1\rvert\leq t$. 

%%%%%%%%%%%%%%%%%%%%%%%%%
%%%% Nec Case %%%%%%
%%%%%%%%%%%%%%%%%%%%%%%%
\section*{Appendix C: Proof of Theorem~\ref{thm:nec} (Necessary condition)}
\label{sec:nec}
We now prove the necessary condition which we have stated in Theorem~\ref{thm:nec}.
Let $d$, $\lambda$, $\epsilon$, and $\sigma$ satisfy the condition of Theorem~\ref{thm:nec}.  
As in the beginning of the proof for Theorem~\ref{thm:suff}, we can deduce the following necessary condition. For all $t\leq1$ it holds that 
\begin{align*}
    \mathbb{P}\left[\min_{i\in[\lambda]}||x^*-x_i||^2 \leq \left(1-\epsilon\right)||x^*||^2\big| \lvert\frac{||x^*||^2}{d}-1\rvert\leq t\right]
   \mathbb{P}\left[\lvert\frac{||x^*||^2}{d}-1\rvert\leq t\right]\\
   +\mathbb{P}\left[\lvert\frac{||x^*||^2}{d}-1\rvert> t\right]\geq \delta
\end{align*}
Then there exists $x^*$ such that $\lvert\frac{||x^*||^2}{d}-1\rvert\leq t$ and 
\begin{eqnarray}
\mathbb{P}\left[\min_{i\in[\lambda]}||x^*-x_i||^2 \leq \left(1-\epsilon\right)||x^*||^2\big| x^*\right]
  \geq\frac{\delta- \mathbb{P}\left[\lvert\frac{||x^*||^2}{d}-1\rvert > t\right]}{ \mathbb{P}\left[\lvert\frac{||x^*||^2}{d}-1\rvert\leq t\right]}\,. 
\label{eq:nec}
\end{eqnarray}
Set $\delta' := \frac{\delta- \mathbb{P}\left[\lvert\frac{||x^*||^2}{d}-1\rvert > t\right]}{ \mathbb{P}\left[\lvert\frac{||x^*||^2}{d}-1\rvert\leq t\right]} $. 
Then the necessary condition~\eqref{eq:nec} can be written as
$$\mathbb{P}\left[Y\leq -\left(\epsilon\frac{||x^*||^2}{\sigma^2}+d\right)\right]\geq1-(1-\delta')^{1/\lambda}$$
with $Y := U-(\frac{||x^*||^2}{\sigma^2}+d)$ and 
$U$ being distributed according to a non-central $\chi^2$ distribution with $d$ degrees of freedom and non-centrality parameter $||x^*||^2/\sigma^2$.
According to the concentration bound~\eqref{concentration2}, we have 
$$\mathbb{P}\left(Y\leq -\left(\epsilon\frac{||x^*||^2}{\sigma^2}+d\right)\right) \leq\exp{\left(-\frac14 \frac{(\epsilon\frac{||x^*||^2}{\sigma^2}+d)^2}{2\frac{||x^*||^2}{\sigma^2}+d}\right)}.$$
%
% From this we derive a necessary condition for 
Condition~\eqref{eq:nec} therefore requires   
$$ \exp{\left(-\frac14 \frac{(\epsilon\frac{||x^*||^2}{\sigma^2}+d)^2}{2\frac{||x^*||^2}{\sigma^2}+d}\right)}\geq1-(1-\delta')^{1/\lambda}.$$
From this we derive
$\epsilon \leq \left( \sqrt{ \tilde{A}_\lambda\left(2\frac{||x^*||^2}{\sigma^2}+d\right)}-d \right)\frac{\sigma^2}{||x^*||^2},$ 
with $\tilde{A}_\lambda =-4\log\left(1-(1-\delta')^{1/\lambda}\right)$. 
As $\epsilon>0$, we obtain that 
$$\sigma^2< \tilde{\sigma}^2 := 2\frac{||x^*||^2/d}{\frac{d}{\bar{A}_\lambda}-1}.$$ 
Fixing $t = d^{-1/3}$ and considering the requirements stated in~\eqref{ass3b} we obtain that 
$\tilde{\sigma}= 2\frac{\tilde{A}_\lambda}{d}+o\left(\frac{\tilde{A}_\lambda}{d}\right)=8 \frac{\log \lambda}{d}+o\left(\frac{\log \lambda}{d}\right)$,
which concludes the proof of the necessary condition, as it shows 
$\sigma^2\in O\left( \frac{\log\lambda_d}{d}\right)$.
\hfill \qed

%%%%%%%%%%%%%%%%%%%%%%%%%
%%%% Upper bound %%%%%%
%%%%%%%%%%%%%%%%%%%%%%%%
\section*{Appendix D: Proof of Theorem~\ref{thm:approx} (Upper Bound for the Gain)}
\label{sec:approx}

The proof of Theorem~\ref{thm:approx} uses the same argument as the one of Theorem~\ref{thm:nec}. We have proved that $\sigma^2$ must be between $0$ and $\tilde{\sigma}=2\frac{||x^*||^2/d}{\frac{d}{\bar{A}_\lambda}-1}$. Then  we get that:

$$\epsilon\leq \sup_{\sigma\in\left[0,\tilde{\sigma}\right]}\left( \sqrt{ \tilde{A}_\lambda\left(2\frac{||x^*||^2}{\sigma^2}+d\right)}-d \right)\frac{\sigma^2}{||x^*||^2}.$$

Noticing that: 
\begin{align*}
    \sup_{\sigma\in\left[0,\tilde{\sigma}\right]}&\left( \sqrt{ \tilde{A}_\lambda\left(2\frac{||x^*||^2}{\sigma^2}+d\right)}-d \right)\frac{\sigma^2}{||x^*||^2} \\
    &= \sup_{\alpha\in\left[0,1\right]}\left( \sqrt{ \tilde{A}_\lambda\left(2\frac{||x^*||^2}{\alpha\tilde{\sigma}^2}+d\right)}-d \right)\frac{\alpha\tilde{\sigma}^2}{||x^*||^2}
\end{align*}

We get after simple algebraic simplifications and for $d$ sufficiently large under assumptions~\eqref{ass3b}:
\begin{align*}
     \sup_{\sigma\in\left[0,\tilde{\sigma}\right]}&\left( \sqrt{ \tilde{A}_\lambda\left(2\frac{||x^*||^2}{\sigma^2}+d\right)}-d \right)\frac{\sigma^2}{||x^*||^2}\\
     &\leq\frac{d\tilde{\sigma}^2}{||x^*||^2} \sup_{\alpha\in\left[0,1\right]}\alpha\left(\sqrt{\alpha^{-1}+\frac{\tilde{A}_\lambda}{d^2}}-1\right)\\
     &\leq\frac{d\tilde{\sigma}^2}{||x^*||^2} \sup_{\alpha\in\left[0,1\right]}\alpha\left(\sqrt{\alpha^{-1}+1}-1\right)\\
     &\leq 8\frac{\log\lambda}{d}+o\left(\frac{\log\lambda}{d}\right)
\end{align*}

Then $\epsilon\in O\left( \frac{\log\lambda_d}{d}\right)$, which concludes the proof of Theorem~\ref{thm:approx}.
\hfill \qed
% \newpage

\end{document}